\pdfoutput=1
\documentclass[11pt]{article}

\usepackage[]{ACL2023}

\usepackage{times}
\usepackage{latexsym}
\usepackage{algorithm}
\usepackage{algorithmic}
\usepackage{booktabs}
\usepackage{times}
\usepackage{latexsym}
\usepackage{algorithm}
\usepackage{algorithmic}
\usepackage{booktabs}
\usepackage{adjustbox}
\usepackage{booktabs}
\usepackage{multirow}
\usepackage{subcaption}
\usepackage{graphicx}
\usepackage{tikz}
\usetikzlibrary{calc}

\usepackage[colorinlistoftodos,prependcaption,textsize=tiny]{todonotes}

\usepackage[T1]{fontenc}
\usepackage[utf8]{inputenc}

\usepackage{microtype}

\usepackage{inconsolata}

\usepackage{amsmath,amsthm,amssymb}
\DeclareMathOperator*{\argmin}{arg\,min}
\DeclareMathOperator*{\argmax}{arg\,max}
\newcommand{\Adv}{\texttt{Adversarial}}
\newcommand{\Uncon}{\texttt{Unconstrained}}
\newcommand{\UnconA}{\texttt{Augmented}}
\newcommand{\Anarchist}{\texttt{Alternate}}
\newcommand{\Abstract}{\texttt{Abstract}}
\newcommand{\FairGrad}{\texttt{FairGrad}}
\newcommand{\FairMixup}{\texttt{Fair MixUp}}
\newcommand{\DF}{\text{DF}}
\newcommand{\INLP}{\texttt{INLP}}
\newcommand{\CDF}{\texttt{DF-Classifier}}
\newcommand{\IFa}[1]{\text{IF}_{#1}}
\usepackage{xifthen}
\usepackage{multirow}
\usepackage{comment}
\usepackage{wrapfig}

\allowdisplaybreaks

\makeatletter
\newcommand\footnoteref[1]{\protected@xdef\@thefnmark{\ref{#1}}\@footnotemark}
\makeatother
\usepackage{array}
\newcolumntype{L}[1]{>{\raggedright\let\newline\\\arraybackslash\hspace{0pt}}m{#1}}
\newcolumntype{C}[1]{>{\centering\let\newline\\\arraybackslash\hspace{0pt}}m{#1}}
\newcolumntype{R}[1]{>{\raggedleft\let\newline\\\arraybackslash\hspace{0pt}}m{#1}}

\mathchardef\mhyphen="2D

\def\tRhelp#1#2\relax{L_{\csname dom#1\endcsname#2}} % True Risk

\def\eRhelp#1#2\relax{\hat{L}_{\csname set#1\endcsname#2}}
\newcommand{\loss}[1]{\ell\ifthenelse{\isempty{#1}{}}{}{\left(#1\right)}}

\newcommand{\mydefalg}[1]{\expandafter\newcommand\csname alg#1\endcsname{\mathcal{#1}}}
\newcommand{\mydefallalg}[1]{\ifx#1\mydefallalg\else\mydefalg{#1}\expandafter\mydefallalg\fi}
\mydefallalg A\mydefallalg % Define a bunch of algorithms accessible with the command \algletter

\newcommand{\mydefdom}[1]{\expandafter\newcommand\csname dom#1\endcsname{\mathcal{#1}}}
\newcommand{\mydefalldom}[1]{\ifx#1\mydefalldom\else\mydefdom{#1}\expandafter\mydefalldom\fi}
\mydefalldom HSTVX\mydefalldom % Define a bunch of domains accessible with the command \domletter

\newcommand{\mydefset}[1]{\expandafter\newcommand\csname set#1\endcsname{\mathcal{#1}}}
\newcommand{\mydefallset}[1]{\ifx#1\mydefallset\else\mydefset{#1}\expandafter\mydefallset\fi}
\mydefallset BCGHPQSTVX\mydefallset % Define a bunch of sets accessible with the command \setletter

\newcommand{\mydefdistr}[1]{\expandafter\newcommand\csname distr#1\endcsname{\mathcal{D}_{\csname dom#1\endcsname}}}
\newcommand{\mydefalldistr}[1]{\ifx#1\mydefalldistr\else\mydefdistr{#1}\expandafter\mydefalldistr\fi}
\mydefalldistr DHSTVXZ\mydefalldistr % Define a bunch of distribution over a domain accessible with the command \distrletter

\newcommand{\mydefspace}[1]{\expandafter\newcommand\csname space#1\endcsname{\mathcal{#1}}}
\newcommand{\mydefallspace}[1]{\ifx#1\mydefallspace\else\mydefspace{#1}\expandafter\mydefallspace\fi}
\mydefallspace ADFGHKLMPRSUVXYZ\mydefallspace % Define a bunch of spaces accessible with the command \spaceletter

\newcommand{\mydeff}[1]{\expandafter\newcommand\csname f#1\endcsname[2][]{#1##1\ifthenelse{\equal{##2}{}}{}{\!\left(##2\right)}}}
\newcommand{\mydefallf}[1]{\ifx#1\mydefallf\else\mydeff{#1}\expandafter\mydefallf\fi}
\mydefallf cdfghkwCFGHMRT{PV}\mydefallf % Define a bunch of functions accessible with the command \fletter

\newcommand{\mydeffsym}[1]{\expandafter\newcommand\csname f#1\endcsname[2][]{\csname #1\endcsname##1\ifthenelse{\equal{##2}{}}{}{\!\left(##2\right)}}}
\newcommand{\mydefallfsym}[1]{\ifx#1\mydefallfsym\else\mydeffsym{#1}\expandafter\mydefallfsym\fi}
\mydefallfsym {phi}{epsilon}{eta}\mydefallfsym % Define a bunch of functions accessible with the command \fletter

\newcommand{\mydefnset}[1]{\expandafter\newcommand\csname nset#1\endcsname{\mathbb{#1}}}
\newcommand{\mydefallnset}[1]{\ifx#1\mydefallnset\else\mydefnset{#1}\expandafter\mydefallnset\fi}
\mydefallnset CNRSZ\mydefallnset % Define a bunch of sets accessible with the command \nsetletter

\newtheorem*{mythm*}{Theorem}

\newtheorem*{mylem*}{Lemma}

\theoremstyle{definition}

\newtheorem*{myrem*}{Remark}

\newtheorem*{myex*}{Example}

\title{Synthetic Data Generation for Intersectional Fairness by Leveraging Hierarchical Group Structure}

\author{Gaurav Maheshwari\textsuperscript{1}, Aur\'elien Bellet\textsuperscript{2}, Pascal Denis\textsuperscript{1}, Mikaela Keller\textsuperscript{1}   \\
\textsuperscript{1} Univ. Lille, Inria, CNRS, Centrale Lille, UMR 9189 - CRIStAL, F-59000 Lille, France\\
\textsuperscript{2} Inria, Université de Montpellier, France\\
  \texttt{first\_name.last\_name@inria.fr} }

\begin{document}
\maketitle
\begin{abstract}
\looseness=-1 In this paper, we introduce a data augmentation approach specifically tailored to enhance intersectional fairness in classification tasks. Our method capitalizes on the hierarchical structure inherent to intersectionality, by viewing groups as intersections of their parent categories. This perspective allows us to augment data for smaller groups by learning a transformation function that combines data from these parent groups. Our empirical analysis, conducted on four diverse datasets including both text and images, reveals that classifiers trained with this data augmentation approach achieve superior intersectional fairness and are more robust to ``leveling down'' when compared to methods optimizing traditional group fairness metrics.
\end{abstract}

\section{Introduction}

The primary objective of fair machine learning is to create models that are free from discriminatory behavior towards subgroups within the population. These subgroups are often defined based on sensitive demographic attributes such as gender (e.g., Male/Female), race (e.g., African-American/European-American), or age (e.g., young/old). To address the above challenge, various strategies have been devised, including pre-processing datasets~\citep{kamiran2012data, feldman2015certifying}, modifying the training process~\citep{cotter2019two, lohaus2020too}, and calibrating outputs of trained models~\citep{iosifidis2019fae, chzhen2019leveraging}. Predominantly, these methods have focused on settings where sensitive groups are identified by a \emph{single} demographic attribute. However, recent studies \citep{DBLP:conf/nips/YangCK20, DBLP:conf/fat/BuolamwiniG18, DBLP:conf/nips/KirkJVIBDSA21} demonstrate that ensuring fairness for an individual attribute does not guarantee \emph{intersectional fairness}, which arises when considering \emph{multiple} attributes concurrently (for example, comparing Male European-Americans or Female African-Americans). For instance, \citet{DBLP:conf/fat/BuolamwiniG18} found that several face recognition systems exhibit significantly higher error rates for darker-skinned females than for lighter-skinned males. These observations are inline with the hypothesis of \citet{Crenshaw1989-CREDTI} that multiple sensitive attributes ``intersect'' to create unique effects.
% multiple sensitive attributes may ``intersect'' to create unique effects. 

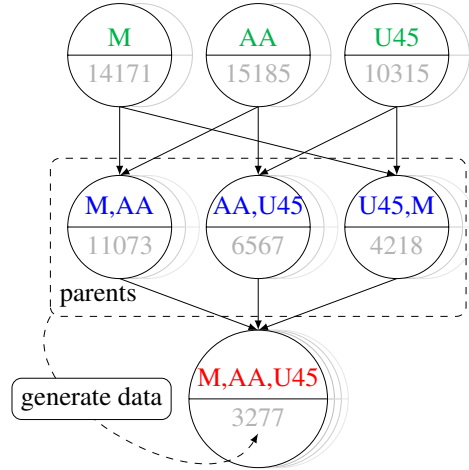
\begin{figure}
    \centering
\resizebox{0.4\textwidth}{!}{
\begin{tikzpicture}

  \coordinate (M) at (1, 4.5);
  \coordinate (AA) at (3, 4.5);
  \coordinate (U45) at (5, 4.5);

  \coordinate (M_AA) at (1, 2);
  \coordinate (AA_U45) at (3, 2);
  \coordinate (U45_M) at (5, 2);

  \coordinate (M_AA_U45) at (3, -0.5);

  % ghost groups 
  \draw[color=gray!50] ($(M_AA_U45)+(0.1,-0)$) circle (1) ++(1,0) -- ++(-2,0);
  \draw[color=gray!40] ($(M_AA_U45)+(0.2,-0)$) circle (1) ++(1,0) -- ++(-2,0);
  \draw[color=gray!20] ($(M_AA_U45)+(0.3,-0)$) circle (1) ++(1,0) -- ++(-2,0);

  \draw[color=gray!40] ($(M_AA)+(0.2,-0)$) circle (0.75)  ++(0.75,0) -- ++(-1.5,0);
  \draw[color=gray!40] ($(AA_U45)+(0.2,-0)$) circle (0.75)  ++(0.75,0) -- ++(-1.5,0);
  \draw[color=gray!40] ($(U45_M)+(0.2,-0)$) circle (0.75)  ++(0.75,0) -- ++(-1.5,0);
  \draw[color=gray!20] ($(M_AA)+(0.4,-0)$) circle (0.75)  ++(0.75,0) -- ++(-1.5,0);
  \draw[color=gray!20] ($(AA_U45)+(0.4,-0)$) circle (0.75)  ++(0.75,0) -- ++(-1.5,0);
  \draw[color=gray!20] ($(U45_M)+(0.4,-0)$) circle (0.75)  ++(0.75,0) -- ++(-1.5,0);

  \draw[color=gray!40] ($(M)+(0.3,-0)$) circle (0.75)  ++(0.75,0) -- ++(-1.5,0);
  \draw[color=gray!40] ($(AA)+(0.3,-0)$) circle (0.75)  ++(0.75,0) -- ++(-1.5,0);
  \draw[color=gray!40] ($(U45)+(0.3,-0)$) circle (0.75)  ++(0.75,0) -- ++(-1.5,0);

  % sensitive group, parents and grand-parents
  \draw [fill=white] (M_AA_U45) circle (1) ++(1,0) -- ++(-2,0);
  \draw [fill=white] (M_AA) circle (0.75)  ++(0.75,0) -- ++(-1.5,0);
  \draw [fill=white] (AA_U45) circle (0.75)  ++(0.75,0) -- ++(-1.5,0);
  \draw [fill=white] (U45_M) circle (0.75)  ++(0.75,0) -- ++(-1.5,0);
  \draw [fill=white] (M) circle (0.75)  ++(0.75,0) -- ++(-1.5,0);
  \draw [fill=white] (AA) circle (0.75)  ++(0.75,0) -- ++(-1.5,0);
  \draw [fill=white] (U45) circle (0.75)  ++(0.75,0) -- ++(-1.5,0);

  % labels
  \draw (M_AA_U45) node[color=red, above] {M,AA,U45};
  \draw (M_AA_U45) node[color=gray!60, below] {3277};
  \draw (M_AA) node[color=blue, above] {M,AA};
  \draw (M_AA) node[color=gray!60, below] {11073};
  \draw (AA_U45) node[color=blue, above] {AA,U45};
  \draw (AA_U45) node[color=gray!60, below] {6567};
  \draw (U45_M) node[color=blue, above] {U45,M};
  \draw (U45_M) node[color=gray!60, below] {4218};
  \draw (M) node[color=blue!30!green, above] {M};
  \draw (M) node[color=gray!60, below] {14171};
  \draw (AA) node[color=blue!30!green, above] {AA};
  \draw (AA) node[color=gray!60, below] {15185};
  \draw (U45) node[color=blue!30!green, above] {U45};
  \draw (U45) node[color=gray!60, below] {10315};
  
  % arrows
  \draw [-latex] ($(M)+(0,-0.75)$) -- ($(U45_M)+(0,0.75)$);
  \draw [-latex] ($(M)+(0,-0.75)$) -- ($(M_AA)+(0,0.75)$);
  \draw [-latex] ($(AA)+(0,-0.75)$) -- ($(AA_U45)+(0,0.75)$);
  \draw [-latex] ($(AA)+(0,-0.75)$) -- ($(M_AA)+(0,0.75)$);  
  \draw [-latex] ($(U45)+(0,-0.75)$) -- ($(U45_M)+(0,0.75)$);
  \draw [-latex] ($(U45)+(0,-0.75)$) -- ($(AA_U45)+(0,0.75)$);
  \draw [-latex] ($(M_AA)+(0,-0.75)$) -- ($(M_AA_U45)+(0,1)$);
  \draw [-latex] ($(AA_U45)+(0,-0.75)$) -- ($(M_AA_U45)+(0,1)$);
  \draw [-latex] ($(U45_M)+(0,-0.75)$) -- ($(M_AA_U45)+(0,1)$);

  \draw [rounded corners, dashed] (0,.7) node[above right]{parents} rectangle (6, 3);%, color=gray!60
  \draw[->,>=latex, dashed] (0,.7)to[bend right=90] ($(M_AA_U45)+(0,-0.5)$) ; % color=gray!60, 
  \draw (1.75, -.5) node[left, draw, rounded corners, fill=white]{generate data}; % color=gray!60,
\end{tikzpicture} }
\caption{Snippet of the hierarchical structure found in intersectional fairness for Twitter Hate Speech Dataset~\citep{huang-etal-2020-multilingual} with 3 sensitive attributes. Here, 'M' stands for Male, 'AA' African American, and 'U45' age under 45 years. The group labeled 'M,AA,U45', represents African American men who are less than 45 years old, and has parent groups 'M,AA', 'M,U45', and 'AA,U45'. For each group, the number of examples is reported. The deeper we go in this hierarchical structure, the smaller the number of examples. Our approach consists in generating additional data for smaller groups by combining data from parent groups.}
    \label{fig:data-gen:hierarchical}
\end{figure}

In response to emerging challenges, there has been a notable shift towards intersectionality in fair machine learning research~\citep{DBLP:journals/corr/abs-2302-12683, DBLP:conf/icde/FouldsIKP20}. Among them, recent studies~\citep{maheshwari-etal-2023-fair, DBLP:conf/cvpr/ZietlowLBKLS022, DBLP:journals/corr/abs-2302-02404} have highlighted that several methods improve intersectional fairness by actually harming the subgroups. In other words, they tend to decrease performance over individual subgroups to achieve better overall fairness, an effect referred to as ``leveling down''. %. This tendency, where an intervention enforces fairness at the expense of the involved groups, is termed ``leveling down." 

In this work, we hypothesize that leveling down can be countered by generating additional data for smaller groups so as to improve their representation. To this end, we propose a data augmentation mechanism that utilizes the hierarchical structure inherent to intersectionality. More precisely, we augment subgroups by modifying and combining data from parent groups (which generally have more examples). Figure~\ref{fig:data-gen:hierarchical} illustrates this hierarchical structure for the Twitter Hate Speech Dataset, showing how the group 'African American, Male, under 45' is composed of 'Male, African American', 'Male, under 45', and 'African American, under 45' groups. It also highlights the data scarcity challenge, showing that the number of samples often decreases sharply as we consider more intersections. For example, the 'African American, Male, under 45' group has 3,277 instances, whereas the 'Male' group has 14,171 instances.

In order to produce valuable examples despite limited data availability, we propose a simple parameterization of the generative model and train it using a loss based on Maximum Mean Discrepancy (MMD)~\cite{DBLP:journals/jmlr/GrettonBRSS12}. This loss quantifies the difference between the original examples from a group and the examples generated by combining examples from its parent groups. Then, we train a classifier on the combination of original and generated examples, using equal sampling~\citep{kamiran2009classifying, gonzalez2021optimising}. The first step increases the diversity of examples the classifier is trained on, thereby improving generalization, while the latter ensures that equal importance is given to all subgroups instead of focusing more on larger groups. We empirically evaluate the quality and diversity of the generated examples and their impact on fairness and accuracy. Our results on various datasets show that our proposed approach  consistently improves fairness, without harming the groups and at a small cost in accuracy.

\section{Related Work}
\label{main:related-work}

In this section, we provide a brief overview of approaches which specifically optimize intersectional fairness. For a more detailed overview, please refer to Appendix~\ref{app:related-work}. ~\citet{DBLP:conf/icde/FouldsIKP20} introduced an in-processing technique that incorporates an intersectional fairness regularizer into the loss function, balancing fairness and accuracy. Conversely,~\citet{DBLP:journals/corr/abs-1911-01468} suggests a post-processing mechanism that adjusts the threshold of the classifier and randomizes predictions for each subgroup independently. InfoFair~\citep{DBLP:conf/bigdataconf/KangXWMT22} adopts a distinct approach by minimizing mutual information between predictions and sensitive attributes. Recently, research has begun to explore the phenomenon of ``leveling down'' in fairness.~\citet{maheshwari-etal-2023-fair, DBLP:journals/corr/abs-2302-02404} argue that the strictly egalitarian perspective of current fairness measures contributes to this phenomenon. Meanwhile,~\citet{DBLP:conf/cvpr/ZietlowLBKLS022} demonstrates leveling down in computer vision contexts and introduces an adaptive augmented sampling strategy using generative adversarial networks~\citep{DBLP:journals/corr/GoodfellowPMXWOCB14} and SMOTE~\citep{DBLP:journals/jair/ChawlaBHK02}. Our work aligns with these developments; however, we propose a modality-independent technique that effectively leverages the intrinsic hierarchical structure of intersectionality.

\section{Problem Statement}
\label{data-gen:notations}
Let $p$ denote the number of distinct \textit{sensitive axes} of interest, which we denote as $\spaceA_{1}, \dots, \spaceA_{p}$. Each of these sensitive axes is a set of discrete-valued \textit{sensitive attributes}. 
% In the following, $\spaceA_{i}=\{0,1\}, \forall i$.
%For instance, a dataset may be composed of gender, race, and age as the three sensitive axes, and each of these sensitive axes may be encoded by a set of sensitive attributes, such as gender: \{male, female\}, race: \{European American, African American\}, and age: \{under $45$, above $45$\}.

Consider a feature space $\spaceX$, a finite discrete label space $\spaceY$%, and the sensitive axis space $\spaceA_{1} \cdots \spaceA_{p}$ corresponding to sensitive axes as defined above. 
. Let $\mathcal{D}$ be an unknown distribution over $\spaceX\times\spaceY\times \spaceA_{1}\times \dots\times \spaceA_{p}$ which can be written as:
% \begin{equation}
%     \mathcal{D} = \sum_{\mathbf{a_1} \in A_1}\cdots\sum_{\mathbf{a_p} \in A_p}P(X,Y,A_1 = \mathbf{a_1}, \cdots, A_p = \mathbf{a_p})
% \end{equation}
\begin{equation}
    \mathcal{D} = P(X,Y,A_1, \cdots, A_p)
\end{equation}

We define a \textit{sensitive group} $\mathbf{g}$ as any $p$-dimensional vector in the Cartesian product set $\spaceG = \spaceA_{1} \times \dots \times \spaceA_{p}$  of the sensitive axes. For instance, a sensitive group $\textbf{g} \in \spaceG$ can be represented as $(a_1, \dots, a_p)$ with corresponding distribution as:
% and the corresponding distribution is given by:
% Thus, $\mathcal{D}_{\mathbf{g}}$, which represents the distribution over group $\mathbf{g}$ can be represented as:
% \begin{equation}    
%     \mathcal{D} = \sum_{\mathbf{g} \in \spaceG} \mathcal{D}_\mathbf{g}
% \end{equation}
% where 
\begin{align*}
    \mathcal{D}_\mathbf{g} &= P(X,Y,A_1 = a_1, \cdots, A_p = a_p) % = P(X,Y,\mathbf{g})\\
\end{align*}

We also introduce a more general group than $\mathbf{g}$ called $\mathbf{g}^{\backslash i}$, referred to as the \textit{parent group} in which the $i$-th sensitive axis is left underspecified. It can be represented as $(a_1, \cdots, a_{i-1}, a_{i+1}, \cdots, a_p)$ where $i \in \{1,\dots,p\}$. The distribution over such a group can be written as:
% Additionally, we also introduce the notion of parents of a group. More specifically, $\mathcal{D}_{\mathbf{g}^{\backslash i}}$ represents an abstract group of $\mathbf{g}$ where $i^{th}$ sensitive axis is not considered, i.e: 
% $P(x,y,a_1, \dots, a_{i-1}, a_{i+1}, \cdots, a_p)$. In other words:
\begin{equation}
\begin{split}    
    \mathcal{D}_{\mathbf{g}^{\backslash i}} &= \textstyle\sum_{\mathit{a_i} \in \spaceA_{i}}P(X,Y,A_1=a_1, \\
    &\:\:\:\:\:\cdots,A_i=\mathit{a_{i}},\cdots, A_p=a_p)
    % &= \sum_{\mathit{a_i} \in \spaceA_{i}}P(X,Y,\mathbf{g})
\end{split}
\end{equation}
\looseness=-1 In our example above, if group $\mathbf{g}$ is \{male, European American, under $45$\}, then the corresponding parent groups are: \{male,  European American\}, \{male,  under $45$\}, \{European American,  under $45$\}.

Finally, in this work, we focus on classification problems and assume $K$ distinct labels. We will denote the distribution of a group conditioned on same label $k$ by $\mathcal{D}_{\mathbf{g}|Y=k}$.

\paragraph{Problem Statement:}
% Consider a feature space $\spaceX$, a finite discrete label space $\spaceY$, and a set $\spaceG$ representing all possible intersections of $p$ sensitive axes as defined above.
As standard in machine learning, $\mathcal{D}$ is generally unknown and instead we have access to a finite dataset $\setT = \{(x_j,y_j,\mathbf{g}_j)\}_{j=1}^n$ consisting of $n$ i.i.d examples sampled from $\mathcal{D}$.
% where $\mathbf{g}_i \in \spaceG,\: \forall i $.
This sample can be rewritten as $\setT =\bigcup_{\mathbf{g}\in\spaceG}\setT_{\mathbf{g}}$ where $\setT_{\mathbf{g}}$ represents the subset of examples from group $\mathbf{g}$. Examples belonging to parent group $\mathbf{g}^{\backslash i}$ are denoted by:
\begin{equation}
    \setT_{\mathbf{g}^{\backslash i}} = \textstyle\bigcup_{\mathit{a_{i}} \in \spaceA_i} \setT_{a_1, \cdots, \mathit{a_i}, \cdots a_p}
\end{equation}

%The goal of fair machine learning is then to learn an accurate model $\fh[_\theta]{} \in \spaceH$, with learnable parameters $\theta \in \nsetR^D$, such that $\fh[_\theta]{} : \spaceX \rightarrow \spaceY$ is fair with respect to a given group fairness definition
The goal of fair machine learning is then to learn an accurate model $\fh{} \in \spaceH$, such that $\fh{} : \spaceX \rightarrow \spaceY$ is fair with respect to a given group fairness definition
% measure
like Equal Opportunity~\citep{DBLP:conf/nips/HardtPNS16}, Equal Odds~\citep{DBLP:conf/nips/HardtPNS16}, Accuracy Parity~\cite{zafar2017fairnessb}, etc.
% Additionally, we also introduce the notion of parents of a group. ore specifically, $\setT_{\mathbf{g}^{\backslash i}}$ represents an set of examples  of where $i^{th}$ sensitive axis is not considered for group $\mathbf{g}$, i.e 

\section{Approach}
\label{data-gen:approach}
In this work, we introduce a novel approach for generating data that leverages the underlying structure of intersectional groups. We begin by highlighting the structural properties of interest, and then present our data generation mechanism. Note that in this work, we treat data as vectors, which allows us to encompass a wide range of modalities including images and text. To convert data into vector representations, we may use pre-trained encoders.

% As a running examples, assume a dataset with 3 sensitive axes, namely gender: \{Male, Female\}, race: \{European American, African American\}, and age: \{Under $45$, Above $45$\}. Following the same encoding mechanism as proposed by~\citet{}, a group of all Female, African American, Above $45$ can be denoted as $(1,1,1)$. While a group of all female, European American can be represented as $(1,0,+)$. Finally, let $\mathcal{D}$ be the unknown distribution through which we sample i.i.d a finite dataset $\setT = \{(x_i,y_i,\mathbf{g}_i)\}_{i=1}^n$ consisting of $n$ examples where $\mathbf{g}_i \in \spaceG\: \forall i $. With a slight abuse of notation, let $\mathcal{D}_\mathbf{g}$ represent the unknown distribution through which we sample  $\setT_{\mathbf{g}}$ which represents the subset of examples from group $\mathbf{g}$. We provide more details about the notation and encoding mechanism in Appendix~\ref{}. 

\subsection{Structure of the Data}
% \paragraph{Structure of the data:} 
Using the notations introduced in the previous section, we make the following simple but crucial observation about the structure of the data:
$$\setT_{\textbf{g}} = \textstyle\bigcap_{i=1}^{p} \setT_{\mathbf{g}^{\backslash i}} \;\; \text{ and } \;\;\setT_{\textbf{g}} \subset \setT_{\mathbf{g}^{\backslash i}}\: \: \forall i \in \{1,\dots,p\}.$$
In other words, the intersection of immediate parent groups constitutes the target group $\textbf{g}$, with each parent group containing more examples than the target group itself. For example, all instances of the group Female African American are also part of both the Female and African American groups. Moreover, the common instances between the Female and African American groups collectively define the Female African American group.

% In other words, the intersection of immediate parent groups results in that group, and there are more examples in each parent than the group itself. For instance, all examples of the group Female African American also belong to group Female as well as African American. Moreover, the common examples in Female and African American groups forms the group Female African American.

\subsection{Data Generation}
\label{data-gen:subsec-data-gen-approach}
% \paragraph{Data Generation:}

Our goal is to learn a generative function $gen_{\theta,k}$ such that, given a dataset $\setT$, a group $\mathbf{g}$, and task label $k$, the generated distribution $Z_{gen}\sim gen_{\theta, k}(\setT,\mathbf{g})$ is similar to the true distribution $\spaceD_{\mathbf{g}|Y=k}$. 
Based on the above observations, we propose to generate examples for group $\textbf{g}$ by combining and transforming the examples from the corresponding parent groups. This can be achieved by appropriate parameterizations of $gen_{\theta,k}$ which we describe next.

% Note that, in this work we consider data to be in vectors enabling us to accommodate various modalities such as images and text. In order to convert data into vector representations we rely on pre-trained models.

% In this work, we treat data as vectors, which allows us to encompass a range of modalities including images and text. In order to convert data into vector representations, we utilize pre-trained models.

\paragraph{Parameterization of the Generative Function:} In this work, we explore the use of two simple choices for the generative function $gen_{\theta,k}(\setT,\mathbf{g})$ that generates an example $Z_{gen} = (X_{gen}, k, \mathbf{g})$ for a given group $\mathbf{g}$ and label $k$.
% , and a more complex generative function with relatively more number of parameters.
The first parameterization is:
% choice for a generative function $gen_{\theta,k}(\setT,\mathbf{g})$ that would generate an example $Z_{gen} = (X_{gen}, k, \mathbf{g})$ for a given group $\mathbf{g}$ and label $k$ is:
% In this work, we explore a simple generative function with less parameters, and a more complex generative function with relatively more number of parameters. A simple generative function which generates example for group $\mathbf{g}$ is:
% We now provide details on the generative function, as well as our mechanism to generate examples from this generative function.
% Note that, in all cases, we sample with replacement. To accommodate various modalities such as images and text, we convert them into vector representations using pre-trained models. 
% An example of a simple generative function which generates examples for group $\mathbf{g}$ can be:
\begin{equation}
\label{data-gen:simple-gen-fn}
    X_{gen} = \textstyle\sum^{p}_{i=1}\lambda_i X_{\mathbf{g}^{\backslash i}}
\end{equation}
with $Z_{\mathbf{g}^{\backslash i}} = (X_{\mathbf{g}^{\backslash i}},k,\mathbf{g}^{\backslash i}) \sim \mathcal{D}_{\mathbf{g}^{\backslash i}|Y=k}$.
% Finally, $Z_{gen} = (X_{gen}, k)$
% Since $\mathcal{D}_{\mathbf{g}^{\backslash i}} $ is unknown we approximate it with the empirical distribution estimated on $\setT_{\mathbf{g}^{\backslash i}}$. 
In the above equation, $\lambda=(\lambda_1,\dots,\lambda_p) \in \mathbb{R}^p$ are the parameters to optimize based on the loss we define below. 
% Note that, in all cases, we sample with replacement.
In other words, we generate data for group $\mathbf{g}$ by forming weighted combinations of examples from its parent groups.

The second parameterization we consider is:
\begin{equation}
\label{data-gen:complex-gen-fn}
    X_{gen} = \textstyle\sum^{p}_{i=1} W \cdot X_{\mathbf{g}^{\backslash i}}^T,
\end{equation}
where $\mathbf{W} \in \mathbb{R}^{d\times d}$ is a diagonal matrix with $d$ parameters where $d$ is the dimension of the encoded inputs. Here, we use a uniform combination of examples from parent groups, but learn weights for the different features of the representation.

Given the limited data available for many groups, we opt to share parameters across them instead of learning specific parameters for each group. This approach, combined with the relatively simple parameterizations of the generative function, serves to reduce the risk of overfitting (recall that in practice we have very limited data for many groups). However, we still learn a separate model for each label, i.e.,  $gen_{\theta, k}(\setT, \mathbf{g}) \:\: \forall k \in K$, to avoid the added complexity of jointly learning $\spaceX \times \spaceY$.
% We leave the effect of jointly learning the features and label for the future work.  

\paragraph{Training the Generative Models:} To train the generative model $gen_{\theta,k}$, we minimize a loss based on Maximum Mean Discrepancy (MMD). MMD is a non-parametric kernel-based divergence used to assess the similarity between distributions by using samples drawn from those distributions \citep{DBLP:journals/jmlr/GrettonBRSS12}. Formally, the MMD between two samples $S=(z_1,\dots,z_m)$ and $S'=(z'_1,\dots,z'_m)$ can be written as
\begin{align*}
    &MMD^2(S, S') = \textstyle\frac{1}{m(m-1)}\big[ \sum_{i} \sum_{j \neq i} k(z_i, z_j)\big. & \\
    & \big. +\textstyle\sum_{i} \sum_{j \neq i} k(z'_i, z'_j)\big] + \frac{1}{m^2}\sum_{i} \sum_{j} k(z_i, z'_j) &
\end{align*}
where $k$ is a reproducing kernel. In this work, we use the radial basis function kernel $k: (z,z') \mapsto \exp(\left \| z-z' \right \|^2 / 2\sigma^2)$ where $\sigma$ is a free parameter. For completeness, more details about MMD are given in Appendix~\ref{app:mmd}

Our loss function is the MMD between the generated samples and the samples from group $\mathbf{g}$, to which we add the MMD between the generated samples and those from its parent groups.\footnote{In our preliminary set of experiments, we found this additional term brought more diversity in the generated examples.} Formally, this can be written as:
\begin{equation}
\label{data-gen:loss-mdd-sample}
\begin{split}
    L_{\mathbf{g},k}(\theta) &= MMD(S_{gen}, S_{\mathbf{g},k}) + \\
      &\:\:\:\:\: \textstyle\sum^{p}_{i=1}MMD(S_{gen}, S_{\mathbf{g}^{\backslash i},k}),
\end{split}
\end{equation}
where $S_{gen}$ is a batch of examples generated from $gen_{\theta,k}$, $S_{\mathbf{g},k}$ and $S_{\mathbf{g}^{\backslash i},k}$ are batches of examples respectively drawn from $\mathcal{D}_{\mathbf{g}|Y=k}$ and $\mathcal{D}_{\mathbf{g}^{\backslash i}|Y=k}$. Since $\mathcal{D}_{\mathbf{g}|Y=k}$ and $\mathcal{D}_{\mathbf{g}^{\backslash i}|Y=k}$ are unknown, we approximate them with the empirical distribution by sampling with replacement from $\setT_{\mathbf{g}|Y=k}$ and $\setT_{\mathbf{g}^{\backslash i}|Y=k}$.
% , including our methodology for generating examples using this function.
% Algorithm~\ref{alg:generator} in
Appendix~\ref{app:algorithm} details the precise training process to learn the generative models.

\paragraph{Training Classifiers on Augmented Data:} After training the generative models $gen_{\theta,k}$, we use them to create additional training data. 
% This process mirrors the training of the function itself. 
Specifically, for a group $\mathbf{g}$, we sample examples from its corresponding parent groups and pass these samples through the generative models as previously described. 
% In Algorithm~\ref{alg:datagenerations}, this procedure includes all steps except for $4$, $8$, and $9$. 
In this way, we can generate additional data for smaller groups that we use to augment the original training dataset, so as to enhance their representation in downstream tasks. As we will see in the next section, this helps to improve the fairness of the classifier.

\paragraph{Alternative formulations:} An alternative approach to learn $gen_{\theta,k}$ involves using a generative adversarial network (GAN)~\citep{DBLP:journals/corr/GoodfellowPMXWOCB14}. In this setup, the adversary aims to differentiate between two distributions, while the encoder strives to mislead the adversary. However, training GANs presents notable challenges~\citep{thanh2020catastrophic, bau2019seeing}, including the risk of mode collapse, the complexity of nested optimization, and substantial computational demands. By contrast, MMD is more straightforward to implement and train,
% providing an exact analytical solution
with significantly less computational burden.
We also note that, while this work primarily employs MMD, our methodology can be adapted to work with other divergences between distributions, such as Sinkhorn Divergences and the Fisher-Rao Distance. We keep the exploration of other choices of divergences
% on data generation (and, subsequently, on fairness)
for future work.

\section{Experiments}
\label{data-gen:experiments}
Our experiments are designed to (i) assess the quality of the data generated by our approach, and (ii) examine the influence of this data on fairness with a focus on avoiding leveling down as well as maximizing the classification performance for the worst-off group. Before presenting results, we start by outlining the datasets, baselines, and fairness metrics we employ. The code base is available here\footnote{Please check the supplementary material. The final version will be released on GitHub with camera ready version.}.

\paragraph{Datasets:} To demonstrate the broad applicability of our proposed approach, we used four diverse datasets varying in size, demographic diversity, and modality, encompassing both text and images. These datasets are: (i) \textit{Twitter Hate Speech}~\citep{huang-etal-2020-multilingual} comprising of tweets annotated with $4$ demographic attributes; (ii) \textit{CelebA}~\citep{DBLP:conf/iccv/LiuLWT15} composed of human face images annotated with various attributes; (iii) \textit{Numeracy}~\citep{DBLP:conf/emnlp/AbbasiDLNSY21} compiles free text responses denoting the numerical comprehension capabilities of individuals; and (iv) \textit{Anxiety}~\citep{DBLP:conf/emnlp/AbbasiDLNSY21}: indicative of a patient's anxiety levels. Experimental setup, splits, and preprocessing are identical to those of~\citet{maheshwari-etal-2023-fair}. Detailed descriptions are available in the Appendix~\ref{app:datasets}.

\paragraph{Methods:} We benchmark against $6$ baselines, encompassing both generative approaches and methods optimizing for intersectional fairness: (i) \textbf{\Uncon}{} solely optimizes model accuracy, ignoring any fairness measure; (ii) \textbf{\Adv}{} adds an adversary~\citep{li-etal-2018-towards} to \textbf{\Uncon}{}, implementing standard adversarial learning approach; (iii) \textbf{\FairGrad}{}~\citep{DBLP:journals/corr/abs-2206-10923} is an in-processing iterative method that adjusts gradients for groups based on fairness levels; (iv) \textbf{\INLP}{}~\citep{DBLP:conf/acl/RavfogelEGTG20} is a post-processing approach that iteratively trains a classifier and then projects the representation on its null space; (v) \textbf{\FairMixup}{}~\citep{DBLP:conf/iclr/ChuangM21} is a generative approach which enforces fairness by forcing the model to have similar predictions on samples generated by interpolating examples belonging to different sensitive groups; (vi) \textbf{DF Classifier}~\citep{DBLP:conf/icde/FouldsIKP20} adds a regularization tailored to improve intersectional fairness. Our approach \textbf{\UnconA{}} is same as \textbf{\Uncon}{}, but trained on data generated via our proposed data generation mechanism.

In all experiments we employ a three-layer fully connected neural network with hidden layers of sizes $128$, $64$, and $32$ as our classifier. Furthermore, we use ReLU as the activation with dropout fixed to $0.5$. Cross-entropy loss is optimized in all cases, employing the Adam optimizer~\citep{DBLP:journals/corr/KingmaB14} with its default parameters. Finally, for text-based datasets we encode the text using bert-base-uncased~\cite{DBLP:conf/naacl/DevlinCLT19} and for images we employ a pre-trained ResNet18\footnote{https://pytorch.org/vision/stable/models.html}~\citep{he2016deep}. Finally, we use equal sampling as shown effective in previous works~\citep{maheshwari-etal-2023-fair, kamiran2009classifying, gonzalez2021optimising}, ensuring equal number of examples for each group. The number of examples, treated as a hyperparameter, spans a spectrum from undersampling to oversampling regime. For more detailed description of hyperparameters and compute infrastructure, please refer to Appendix~\ref{app:hyperparameters}.

To generate data for \UnconA{}, we employ the generative function as described in Section~\ref{data-gen:subsec-data-gen-approach}. More specifically, our initial experiments suggest that employing a simpler model with fewer parameters (Equation~\ref{data-gen:simple-gen-fn}) for the positive class, and a more complex model with a larger number of parameters for the negative class (Equation~\ref{data-gen:complex-gen-fn}), leads to an enhanced fairness-accuracy trade-off, when using the False Positive rate as fairness measure. Consequently, for the positive class, we implement the function detailed in Equation~\ref{data-gen:simple-gen-fn}, and for the negative class, we apply the model specified in Equation~\ref{data-gen:complex-gen-fn}.

\paragraph{Fairness Metrics:} To assess unfairness, we utilize two fairness definitions specifically designed for the context of intersectional fairness: $\alpha$-Intersectional Fairness ($\IFa{\alpha})$~\citep{maheshwari-etal-2023-fair} and Differential Fairness ($\DF$)~\citep{DBLP:conf/icde/FouldsIKP20}. Detailed descriptions of these metrics are provided in the Appendix (Section~\ref{app:fairness-measure}).

For the performance measure $m$ associated with these definitions, we focus on False Positive Rate. Formally, for a group $\mathbf{g}$, $m$ is given by:
\begin{align*}
    m(h_{\theta}, \setT_{\mathbf{g}}) = 1 - P(h_{\theta}(x)=0|(x,y) \in \setT_{\mathbf{g}}, y=1)
    % \: \forall (x,y)\: \in \setT_{\mathbf{g}}
\end{align*}

To estimate these empirical probabilities, we adopt the bootstrap estimation method proposed by~\citet{DBLP:journals/corr/abs-1911-01468}. We generate $1000$ datasets by sampling from the original dataset with replacement. We then estimate the probabilities on this dataset using a smoothed empirical estimation mechanism and then average the results over all the sampled datasets. In addition to these fairness metrics, we report the performance measure for both the best and worst-performing groups.

\paragraph{Utility metric:} In order to evaluate the utility of various methods, we employ balanced accuracy.

\subsection{Quality of Generated Data}
In this experiment, we assess the quality and diversity of data generated by our approach. Our goal is to generate data that resemble the overall distribution of real data, while ensuring the generated examples remain distinct from the original samples. To this end, we propose two evaluations:

\begin{itemize} 
    \item \textbf{Diversity:} for each generated example, we identify the most similar example in the real dataset. If the generated sample closely resembles a real one, the distance between the generated and real examples will be substantially smaller than between distinct real examples.
    
    \item \textbf{Distinguishability:} we train a classifier to differentiate between generated and real datasets. If the classifier's accuracy approaches that of a random guess, it suggests the empirical distributions of the generated and real data are similar.
\end{itemize}

In both experiments, we report metrics based on the entire dataset rather than computing averages for each group and then aggregating averages.

\subsubsection{Diversity}
In this experiment, we use cosine similarity as a measure of closeness.

We generate $1000$ examples and randomly select an equivalent number from the actual (real) dataset. For each real example, we find its nearest counterpart within the actual dataset to establish a baseline, termed 'R-R'. Then, for every generated example, we identify the closest match in the actual dataset, referred to as 'G-R'. To further assess diversity, we also present results of the closest match of each generated sample in the generated dataset, called 'G-G'. The results of this experiment are presented in Table~\ref{tab:distance}.

Across all datasets, we observe that the distance between generated and real examples is similar to the distance observed between two real examples. In each dataset, the closeness between G-R pairs is less than that observed in R-R pairs. Moreover, the G-G pairs exhibit lower similarity scores compared to R-R pairs, suggesting greater diversity in the generated dataset. Based on these results, we conclude that the generated examples are diverse and not mere replicas of the real samples.

\begin{table}[]
\centering
\begin{tabular}{@{}llll@{}}
\toprule
Dataset             & G-R & R-R & G-G    \\ \midrule
CelebA              &     0.46      &    0.48             &    0.44                 \\
Numeracy            &     0.51      &    0.58             &    0.45                 \\
Anxiety             &     0.51      &    0.59             &    0.46                 \\
Twitter Hate Speech &     0.47      &    0.53             &    0.45                 \\ \bottomrule
\end{tabular}
\caption{Analyzing the similarity of a generated sample with existing sample. For clarity and ease of readability, we have omitted the standard deviation in our reporting, as it remained below $\pm$ 0.01 across all settings.}
\label{tab:distance}
\end{table}

\subsubsection{Distinguishability}
We frame distinguishability as a binary classification task where we train a two-layer MLP classifier aimed at distinguishing between real and generated samples. Again, we compile a dataset by selecting $1000$ real instances and $1000$ generated samples. This dataset is subsequently partitioned into training and test sets with a ratio of 80\% to 20\%.

\begin{table}[]
\centering
\begin{tabular}{@{}ll@{}}
\toprule
Dataset             & Accuracy \\ \midrule
CelebA              &    0.52 $\pm$ 0.011 \\
Numeracy            &    0.64 $\pm$ 0.012 \\
Anxiety             &    0.64 $\pm$ 0.019 \\
Twitter Hate Speech &    0.57 $\pm$ 0.022 \\ \bottomrule
\end{tabular}
\caption{Accuracy of a classifier to distinguish between real and generated sample over various datasets. The value of $0.5$ represents a random classifier, while $1.0$ is a perfect classifier. }
\label{tab:distinguishability}
\end{table}

Results are presented in Table~\ref{tab:distinguishability}. The  mean accuracy of the classifier is approximately $0.59$, suggesting that the generated samples have a distribution similar, but not identical to, the real instances. In our preliminary experiments we found that by modulating the generator complexity (i.e by employing more complex models with more parameters), we could achieve near-random distinguishability. However, such adjustments led to an unfavorable fairness-accuracy trade-off. We conjecture this may arise because near-random indistinguishability in the generated samples causes them to inherit biases from the real data.

\begin{table*}[h!]
    \centering

    \begin{subtable}{\linewidth}
    \vspace*{0.5 cm}
        \centering
        \adjustbox{max width=\linewidth}{
        \begin{tabular}{@{}lccccc@{}}
            \toprule
                Method          & BA $\uparrow$& Best Off $\downarrow$ & Worst Off $\downarrow$   & \DF{} $\downarrow$      & $\IFa{0.5}$   $\downarrow$\\ \midrule
                \Uncon          & 0.63 + 0.01  & 0.25 + 0.02 & 0.51 + 0.03 & 0.43 +/- 0.09 & 0.52 +/- 0.03\\
                \Adv            & 0.63 + 0.01  & 0.27 + 0.06 & 0.55 + 0.12 & 0.48 +/- 0.05 & 0.55 +/- 0.04 \\
                \FairGrad       & 0.63 + 0.01  & 0.29 + 0.05 & 0.56 + 0.12 & 0.48 +/- 0.07 & 0.57 +/- 0.04 \\
                \INLP           & 0.63 + 0.01  & 0.22 + 0.02 & 0.49 + 0.03 & 0.42 +/- 0.07 & 0.48 +/- 0.03\\
                \FairMixup      & 0.61 + 0.01  & 0.28 + 0.02 & 0.55 + 0.06 & 0.47 +/- 0.09 & 0.55 +/- 0.02 \\
                \CDF            & 0.63 + 0.01  & 0.29 + 0.08 & 0.56 + 0.09 & 0.48 +/- 0.17 & 0.56 +/- 0.08 \\
                \UnconA         & 0.6 + 0.0  & 0.13 + 0.08 & 0.35 + 0.12 & 0.29 +/- 0.32 & 0.39 +/- 0.11 \\\bottomrule

        \end{tabular}}
        \caption{\label{main-tab:anxiety} Results on Anxiety}
    \end{subtable}

    \begin{subtable}{\linewidth}
    \vspace*{0.5 cm}
        \centering
        \adjustbox{max width=\linewidth}{
        \begin{tabular}{@{}lccccc@{}}
            \toprule
                Method          & BA $\uparrow$& Best Off $\downarrow$ & Worst Off $\downarrow$   & \DF{} $\downarrow$      & $\IFa{0.5}$   $\downarrow$\\ \midrule
                \Uncon          & 0.81 + 0.0  & 0.18 + 0.01 & 0.46 + 0.01 & 0.42 +/- 0.05  & 0.46 +/- 0.02\\
                \Adv            & 0.79 + 0.01  & 0.18 + 0.01 & 0.48 + 0.04 & 0.46 +/- 0.08 & 0.47 +/- 0.02 \\
                \FairGrad       & 0.8 + 0.0  & 0.17 + 0.01 & 0.49 + 0.03 & 0.49 +/- 0.1    & 0.44 +/- 0.02 \\
                \INLP           & 0.66 + 0.0  & 0.08 + 0.02 & 0.26 + 0.02 & 0.22 +/- 0.25  & 0.29 +/- 0.04\\
                \FairMixup      & 0.81 + 0.01  & 0.18 + 0.02 & 0.46 + 0.02 & 0.42 +/- 0.09 & 0.45 +/- 0.04 \\
                \CDF            & 0.81 + 0.0  & 0.13 + 0.01 & 0.45 + 0.02 & 0.46 +/- 0.1   & 0.39 +/- 0.03 \\
                \UnconA         & 0.81 + 0.0  & 0.06 + 0.01 & 0.36 + 0.03 & 0.38 +/- 0.13  & 0.27 +/- 0.02 \\\bottomrule

        \end{tabular}}
        \caption{\label{main-tab:twitter_hate_speech} Results on Twitter Hate Speech}
    \end{subtable}

    \caption{Test results on (a) \textit{Anxiety}, and (b) \textit{Twitter Hate Speech}. We select hyperparameters based on $\IFa{0.5}$ value. The utility of various approaches is measured by balanced accuracy (BA), whereas fairness is measured by differential fairness ($\DF$) and intersectional fairness ($\IFa{0.5}$) on the False Positive Rate (FPR). For both fairness definitions, lower is better, while for balanced accuracy, higher is better. Best Off and Worst Off represent the min FPR and max FPR across groups (in both cases, lower is better). Results have been averaged over 5 different runs.}
    \label{tab:fairness-accuracy-main}
    
\end{table*}

\subsection{Fairness-Accuracy Trade-offs}

In this experiment, we explore the impact of generated data on the fairness-privacy trade-off and compare our approach to existing fairness-promoting methods.
% To that end, we train \Uncon{} just over the generated data which we refer to as \UnconA{} in this work.
We pay particular attention to the leveling down phenomenon: a method is considered to exhibit leveling down if its performance for the worst-off or best-off group is inferior to that of the unconstrained model. 

The outcomes of this experiment is presented in Table~\ref{tab:fairness-accuracy-main}. Detailed results for CelebA and Numeracy, both of which display a similar trend, are provided in Appendix~\ref{app:extended-results}. In terms of  accuracy, \UnconA{} exhibits a slight drop for the Anxiety dataset. However, its accuracy is on par with the Unconstrained model when evaluated on Twitter Hate Speech. In terms of performance for both best-off and worst-off groups, \UnconA{} outperforms competing methods. Notably, \UnconA{} does not show any signs of leveling down across all datasets. When assessing $\IFa{\alpha}$ with $\alpha=0.5$, \UnconA{} consistently achieves the best fairness results among the datasets. We also plot the complete trade-off between relative and absolute performance of groups by varying $\alpha$ in Figure~\ref{app:fig:overall-tradeoff} in Appendix~\ref{app:extended-results}. For the Anxiety dataset, \UnconA{} gives the best trade-off for every value of $\alpha$. In the case of Twitter Hate Speech, \INLP{} achieves comparable results, although with a noticeable drop in accuracy ($14$ points below \UnconA).
Overall, the results show that our \UnconA{} gives a superior accuracy-fairness trade-off and successfully avoids leveling down.

\subsection{Impact of Intersectionality}
\label{sec:impact-of-intersectionality}

\begin{figure}[t]
    \centering
    \includegraphics[width=\linewidth]{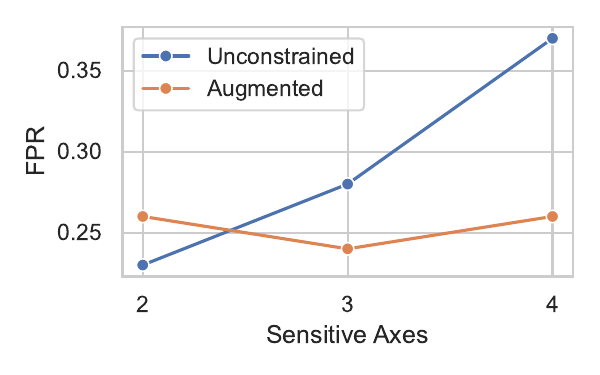}
    \caption{FPR of worst-off group on \textit{CelebA} (the lower, the better) by varying the number of sensitive axes.}
    \label{fig:sensitive-axis-data-gen}
\end{figure}

In this experiment, we examine the influence of intersectionality on our approach and its effect on worst-case performance. To this end, we iteratively introduce more sensitive axes and plot the worst case performance. For example, akin to the experiment in~\citep{maheshwari-etal-2023-fair} using CelebA, we initially consider gender % (selected at random)
as a single sensitive axis. In the subsequent step, we incorporate age % (also selected randomly)
alongside gender. Similarly, we then add attractiveness, and finally skin color. 

The results of this experiment can be found in Figure~\ref{fig:sensitive-axis-data-gen}. With fewer groups (2 sensitive axes), the model's performance on the generated dataset closely matches that on the real dataset. However, as the number of axes increases, the performance difference becomes more pronounced. Furthermore, we find that the performance of the model remains relatively stable despite the increase in sensitive axes, further underscoring the effectiveness of our proposed approach.

\subsection{Alternative Structures}
\label{sec:alternate_structure}

\looseness=-1 Our proposed approach generates additional data for a target group with data from its corresponding parent groups. In this experiment, we explore alternative structures. Taking the target group $\mathbf{g}$ composed of \{male, European American, under $45$\} as an example, we examine two distinct structures:
\begin{itemize}
    \item \Anarchist: Here, we use examples from parent groups unrelated to the target group. More specifically, we follow an adversarial approach where we choose parents such that they share no examples with the target group. For instance, for group $\mathbf{g}$ , we define the adversarial group $\mathbf{\neg g}$ as \{Female, African American, above $45$\}. We then draw examples from parents of group $\mathbf{\neg g}$ for training our generative model for $\mathbf{g}$. We provide the exact formalism and setup in Appendix~\ref{app:sec:hierarchical_structure}.

    \item \Abstract: Here, we use examples from the parents of parents of the target group. For example, for $\mathbf{g}$, the immediate parent groups are: (\{male, European American\}, \{male, under $45$\}, \{European American, under $45$\}). Instead of drawing examples from these immediate parent groups, we use examples from the parents of these parent groups, namely (\{male\}, \{European American\}, \{under $45$\}).
    % In this experiment, we explore the impact of generating data from the parents of immediate parents, as opposed to solely from the immediate parent set
    
\end{itemize}

\begin{table}[h!]
    \centering

    \begin{subtable}{\linewidth}
    \vspace*{0.5 cm}
        \centering
        \adjustbox{max width=\linewidth}{
        \begin{tabular}{@{}lccc@{}}
            \toprule
                Method                      & BA $\uparrow$ & $\IFa{0.5}$   $\downarrow$\\ \midrule
                \Uncon                      & 0.63          & 0.52   \\
                \UnconA                     & 0.60          & 0.39   \\
                \Anarchist                  & 0.61          & 0.40  \\
                \Abstract                   & 0.59          & 0.43 \\ \bottomrule
        \end{tabular}}
        \caption{\label{main-tab:anxiety-something} Results on Anxiety}
    \end{subtable}

    \begin{subtable}{\linewidth}
    \vspace*{0.5 cm}
        \centering
        \adjustbox{max width=\linewidth}{
        \begin{tabular}{@{}lccc@{}}
            \toprule
                Method                      & BA $\uparrow$ & $\IFa{0.5}$   $\downarrow$\\ \midrule
                \Uncon                      & 0.81          & 0.46   \\
                \UnconA                     & 0.81          & 0.27   \\
                \Anarchist                  & 0.81          & 0.29  \\
                \Abstract                   & 0.81          & 0.32 \\ \bottomrule
        \end{tabular}}
        \caption{\label{main-tab:twitter_hate_speech-something} Results on Twitter Hate Speech}
    \end{subtable}

    \caption{Test results on (a) \textit{Anxiety}, and (b) \textit{Twitter Hate Speech} using False Positive Rate showcasing Balanced Accuracy (BA) and $\IFa{0.5}$}
    \label{tab:fairness-accuracy-main-randomized}
    
\end{table}

\begin{figure}
    \centering
    \includegraphics[width=\linewidth]{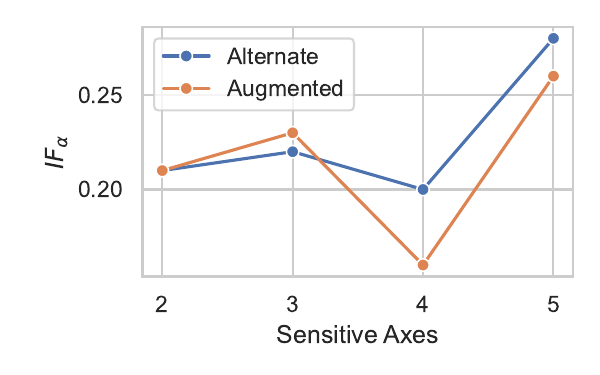}
    \caption{$\IFa{0.5}$ comparison between \UnconA{} and \Anarchist{} by varying the number of sensitive axes on CelebA. With a smaller number of sensitive axes, \Uncon{} and \Anarchist{} exhibit comparable performance. However, as the number of sensitive axes increases, \UnconA{} begins to outperform \Anarchist.
    }
    \label{fig:sensitive-axis-data-gen-randomized}
\end{figure}

The results of these experiments are provided in Table~\ref{tab:fairness-accuracy-main-randomized}. For both experiments, we find that any form of data augmentation approach including the \Anarchist{} improves fairness. For instance, on Anxiety, \Anarchist{} significantly outperforms \Uncon{} and reaches the same level of fairness as \UnconA. Similarly, the \Abstract{} approach outperforms \Uncon{} on Anxiety. These observations indicate that data augmentation via combining from different groups is a viable strategy in general. However, when comparing \Abstract{} performance with~\UnconA{}, we find that \UnconA{} generally outperforms \Abstract. We hypothesize that this occurs because considering more abstract groups approximates a scenario where no groups are considered, which is similar to an unconstrained. % It also increases the variance of examples the generative function sees at both the time of trainign and sampling. Thus decreasib the quakuty of the generated example.

% In our study, we observed that for Twitter-based hate speech and anxiety topics, the anarchist method performs on par with, or even surpasses, the hierarchical approach. This leads us to hypothesize that the effectiveness of a hierarchical structure becomes more pronounced as the number of sensitive axes increases, due to its superior inductive bias. To test this hypothesis, we conducted an experiment akin to that in Section~\ref{sec:impact-of-intersectionality}, gradually increasing the number of sensitive axes. The findings, illustrated in Figure~\ref{fig:sensitive-axis-data-gen-randomized}, indicate that with a limited number of sensitive axes, both approaches yield comparable results. However, as the number of axes expands, the hierarchical approach generally outperforms the anarchist method. It's important to note that these results might also be influenced by inherent dataset characteristics, such as modality, size, and diversity. A comprehensive exploration of how these characteristics interact with the optimal structure for generating fair data is a promising avenue for future research. In conclusion, our experiments suggest that data augmentation across groups is a viable strategy for enhancing fairness in machine learning models.

\looseness=-1 Interestingly, we find that for Twitter Hate Speech and Anxiety, \Anarchist{} performs similarly to \UnconA{} (with a small advantage to the latter in terms of fairness). We hypothesize the hierarchical structure leveraged in \UnconA{} becomes more relevant with the increase in the number of sensitive axes as it provides better inductive bias. To test this hypothesis, we conducted an experiment akin to that in Section~\ref{sec:impact-of-intersectionality} where we gradually increase the number of sensitive axes in CelebA. The findings, illustrated in Figure~\ref{fig:sensitive-axis-data-gen-randomized}, indicate that with a limited number of sensitive axes, both approaches yield comparable results. However, as the number of axes increases, \UnconA{} generally outperforms the \Anarchist{}. It is important to note that these results might also be influenced by inherent dataset characteristics, such as modality, size, and diversity. A comprehensive exploration of how these characteristics interact with the optimal structure for generating augmented data is an interesting avenue for future research. In summary, our experiments show that data augmentation across groups is a viable strategy for enhancing the fairness of machine learning models in intersectional scenarios.

% The results of this experiment are plotted in Figure~\ref{fig:sensitive-axis-data-gen-randomized}. In this experiment we find that, with fewer sensitive axes, the performance between the two approaches is similar, while for more axes, the hierarchical approach tends to outperform the Anarchist approach. However, it might also be the inherent characteristics of the dataset such as the modality, size, and diversity. We leave this investigation of relationship between dataset characteristics and the optimum structure to generate data for fairness to the future. In summary, through these experiments we find that augmenting groups, by transforming data from other groups is a promising approach.

% We hypothesize this to be the case, as these datasets have fewer sensitive axes, and thus, the groups representation are closer.

% We leave the investigation of influence of characteristic of data and the corresponding structure to the future.

\section{Conclusion}
\label{data-gen:conclusion}

In this paper, we introduce a data augmentation mechanism that leverages the hierarchical structure inherent to intersectional fairness. Our extensive experiments demonstrate that this method not only generates diverse data but also enhances the classifier's performance across both the best-off and worst-off  groups. In the future, we plan to extend our approach to a broader range of performance metrics, delve into zero-shot fairness, and explore more sophisticated sampling mechanisms.

\section{Limitations}
\label{sec:limitations}
% While appealing, our proposed data generation mechanism also has limitations. The primary amongst them is that it assumes correct sensitive annotations are available for each data point. Incorrect or missing annotation could results in situations where a perfectly fair model might lead to harm to groups with missing or incorrect annotation. Moreover, it assumes a static view of the fairness problem, and does not take into account problems such as data drift which could lead to unfair model over time. Finally, even though our experiments suggest superior performance, our evaluation is still limited to  datasets and setting we evaluated over. We recommend practitioners using this approach to rigorously evaluate the model while keeping their application specific setting in mind.
While appealing, our proposed data generation mechanism is not without limitations. Its primary constraint is the assumption of accurate sensitive annotations for each data point. Inaccurate or missing annotations could lead to scenarios where an otherwise fair model inadvertently harms groups with incorrect or missing annotations. Additionally, this mechanism adopts a static view of fairness, failing to account for issues like data drift, which may result in the model becoming unfair over time. Furthermore, despite our experiments indicating superior performance, our evaluation is confined to the specific datasets and settings we tested. We advise practitioners employing this approach to conduct thorough evaluations of the model, considering the unique aspects of their intended application.

\section{Acknowledgement}
The authors would like to thank the Agence Nationale de la Recherche for funding this work under grant number ANR-19-CE23-0022, as well as the reviewers for their feedback and suggestions.

\bibliography{anthology,custom}

\begin{thebibliography}{49}
\expandafter\ifx\csname natexlab\endcsname\relax\def\natexlab#1{#1}\fi

\bibitem[{Abbasi et~al.(2021)Abbasi, Dobolyi, Lalor, Netemeyer, Smith, and
  Yang}]{DBLP:conf/emnlp/AbbasiDLNSY21}
Ahmed Abbasi, David~G. Dobolyi, John~P. Lalor, Richard~G. Netemeyer, Kendall
  Smith, and Yi~Yang. 2021.
\newblock \href {https://doi.org/10.18653/v1/2021.emnlp-main.304} {Constructing
  a psychometric testbed for fair natural language processing}.
\newblock In \emph{Proceedings of the 2021 Conference on Empirical Methods in
  Natural Language Processing, {EMNLP} 2021, Virtual Event / Punta Cana,
  Dominican Republic, 7-11 November, 2021}, pages 3748--3758. Association for
  Computational Linguistics.

\bibitem[{Barocas et~al.(2017)Barocas, Bradley, Honavar, and
  Provost}]{barocas2017big}
Solon Barocas, Elizabeth Bradley, Vasant Honavar, and Foster Provost. 2017.
\newblock Big data, data science, and civil rights.
\newblock \emph{arXiv preprint arXiv:1706.03102}.

\bibitem[{Bau et~al.(2019)Bau, Zhu, Wulff, Peebles, Strobelt, Zhou, and
  Torralba}]{bau2019seeing}
David Bau, Jun-Yan Zhu, Jonas Wulff, William Peebles, Hendrik Strobelt, Bolei
  Zhou, and Antonio Torralba. 2019.
\newblock Seeing what a gan cannot generate.
\newblock In \emph{Proceedings of the IEEE/CVF International Conference on
  Computer Vision}, pages 4502--4511.

\bibitem[{Buchanan(2012)}]{buchanan2012ethical}
Elizabeth Buchanan. 2012.
\newblock Ethical decision-making and internet research.
\newblock \emph{Association of Internet Researchers}.

\bibitem[{Buolamwini and Gebru(2018)}]{DBLP:conf/fat/BuolamwiniG18}
Joy Buolamwini and Timnit Gebru. 2018.
\newblock \href {http://proceedings.mlr.press/v81/buolamwini18a.html} {Gender
  shades: Intersectional accuracy disparities in commercial gender
  classification}.
\newblock In \emph{Conference on Fairness, Accountability and Transparency,
  {FAT} 2018, 23-24 February 2018, New York, NY, {USA}}, volume~81 of
  \emph{Proceedings of Machine Learning Research}, pages 77--91. {PMLR}.

\bibitem[{Burrell(2016)}]{burrell2016machine}
Jenna Burrell. 2016.
\newblock How the machine ‘thinks’: Understanding opacity in machine
  learning algorithms.
\newblock \emph{Big data \& society}, 3(1):2053951715622512.

\bibitem[{Calders and Verwer(2010)}]{calders2010three}
Toon Calders and Sicco Verwer. 2010.
\newblock Three naive bayes approaches for discrimination-free classification.
\newblock \emph{Data mining and knowledge discovery}, 21(2):277--292.

\bibitem[{Calmon et~al.(2017)Calmon, Wei, Vinzamuri, Ramamurthy, and
  Varshney}]{calmon2017optimized}
Flavio~P Calmon, Dennis Wei, Bhanukiran Vinzamuri, Karthikeyan~Natesan
  Ramamurthy, and Kush~R Varshney. 2017.
\newblock Optimized pre-processing for discrimination prevention.
\newblock volume~30.

\bibitem[{Chawla et~al.(2002)Chawla, Bowyer, Hall, and
  Kegelmeyer}]{DBLP:journals/jair/ChawlaBHK02}
Nitesh~V. Chawla, Kevin~W. Bowyer, Lawrence~O. Hall, and W.~Philip Kegelmeyer.
  2002.
\newblock \href {https://doi.org/10.1613/JAIR.953} {{SMOTE:} synthetic minority
  over-sampling technique}.
\newblock \emph{J. Artif. Intell. Res.}, 16:321--357.

\bibitem[{Chuang and Mroueh(2021)}]{DBLP:conf/iclr/ChuangM21}
Ching{-}Yao Chuang and Youssef Mroueh. 2021.
\newblock \href {https://openreview.net/forum?id=DNl5s5BXeBn} {Fair mixup:
  Fairness via interpolation}.
\newblock In \emph{9th International Conference on Learning Representations,
  {ICLR} 2021, Virtual Event, Austria, May 3-7, 2021}. OpenReview.net.

\bibitem[{Chzhen et~al.(2019)Chzhen, Denis, Hebiri, Oneto, and
  Pontil}]{chzhen2019leveraging}
Evgenii Chzhen, Christophe Denis, Mohamed Hebiri, Luca Oneto, and Massimiliano
  Pontil. 2019.
\newblock Leveraging labeled and unlabeled data for consistent fair binary
  classification.
\newblock \emph{arXiv preprint arXiv:1906.05082}.

\bibitem[{Commission(2018)}]{european2018communication}
European Commission. 2018.
\newblock Communication artificial intelligence for europe.

\bibitem[{Cotter et~al.(2019)Cotter, Jiang, and Sridharan}]{cotter2019two}
Andrew Cotter, Heinrich Jiang, and Karthik Sridharan. 2019.
\newblock Two-player games for efficient non-convex constrained optimization.
\newblock In \emph{Algorithmic Learning Theory}, pages 300--332. PMLR.

\bibitem[{Crenshaw(1989)}]{Crenshaw1989-CREDTI}
Kimberle Crenshaw. 1989.
\newblock Demarginalizing the intersection of race and sex: A black feminist
  critique of antidiscrimination doctrine, feminist theory and antiracist
  politics.
\newblock \emph{The University of Chicago Legal Forum}, 140:139--167.

\bibitem[{Devlin et~al.(2019)Devlin, Chang, Lee, and
  Toutanova}]{DBLP:conf/naacl/DevlinCLT19}
Jacob Devlin, Ming{-}Wei Chang, Kenton Lee, and Kristina Toutanova. 2019.
\newblock \href {https://doi.org/10.18653/v1/n19-1423} {{BERT:} pre-training of
  deep bidirectional transformers for language understanding}.
\newblock In \emph{Proceedings of the 2019 Conference of the North American
  Chapter of the Association for Computational Linguistics: Human Language
  Technologies, {NAACL-HLT} 2019, Minneapolis, MN, USA, June 2-7, 2019, Volume
  1 (Long and Short Papers)}, pages 4171--4186. Association for Computational
  Linguistics.

\bibitem[{Feldman et~al.(2015)Feldman, Friedler, Moeller, Scheidegger, and
  Venkatasubramanian}]{feldman2015certifying}
Michael Feldman, Sorelle~A Friedler, John Moeller, Carlos Scheidegger, and
  Suresh Venkatasubramanian. 2015.
\newblock Certifying and removing disparate impact.
\newblock In \emph{proceedings of the 21th ACM SIGKDD international conference
  on knowledge discovery and data mining}, pages 259--268.

\bibitem[{Filippi et~al.(2023)Filippi, Zannone, and
  Koshiyama}]{DBLP:journals/corr/abs-2302-12683}
Giulio Filippi, Sara Zannone, and Adriano~S. Koshiyama. 2023.
\newblock \href {https://doi.org/10.48550/ARXIV.2302.12683} {Intersectional
  fairness: {A} fractal approach}.
\newblock \emph{CoRR}, abs/2302.12683.

\bibitem[{Foulds et~al.(2020)Foulds, Islam, Keya, and
  Pan}]{DBLP:conf/icde/FouldsIKP20}
James~R. Foulds, Rashidul Islam, Kamrun~Naher Keya, and Shimei Pan. 2020.
\newblock \href {https://doi.org/10.1109/ICDE48307.2020.00203} {An
  intersectional definition of fairness}.
\newblock In \emph{36th {IEEE} International Conference on Data Engineering,
  {ICDE} 2020, Dallas, TX, USA, April 20-24, 2020}, pages 1918--1921. {IEEE}.

\bibitem[{Gonz{\'a}lez-Zelaya et~al.(2021)Gonz{\'a}lez-Zelaya, Salas, Prangle,
  and Missier}]{gonzalez2021optimising}
Vladimiro Gonz{\'a}lez-Zelaya, Juli{\'a}n Salas, Dennis Prangle, and Paolo
  Missier. 2021.
\newblock Optimising fairness through parametrised data sampling.
\newblock In \emph{EDBT}, pages 445--450.

\bibitem[{Goodfellow et~al.(2014)Goodfellow, Pouget{-}Abadie, Mirza, Xu,
  Warde{-}Farley, Ozair, Courville, and
  Bengio}]{DBLP:journals/corr/GoodfellowPMXWOCB14}
Ian~J. Goodfellow, Jean Pouget{-}Abadie, Mehdi Mirza, Bing Xu, David
  Warde{-}Farley, Sherjil Ozair, Aaron~C. Courville, and Yoshua Bengio. 2014.
\newblock \href {http://arxiv.org/abs/1406.2661} {Generative adversarial
  networks}.
\newblock \emph{CoRR}, abs/1406.2661.

\bibitem[{Gretton et~al.(2012)Gretton, Borgwardt, Rasch, Sch{\"{o}}lkopf, and
  Smola}]{DBLP:journals/jmlr/GrettonBRSS12}
Arthur Gretton, Karsten~M. Borgwardt, Malte~J. Rasch, Bernhard Sch{\"{o}}lkopf,
  and Alexander~J. Smola. 2012.
\newblock \href {https://doi.org/10.5555/2503308.2188410} {A kernel two-sample
  test}.
\newblock \emph{J. Mach. Learn. Res.}, 13:723--773.

\bibitem[{Hardt et~al.(2016)Hardt, Price, and
  Srebro}]{DBLP:conf/nips/HardtPNS16}
Moritz Hardt, Eric Price, and Nati Srebro. 2016.
\newblock \href
  {https://proceedings.neurips.cc/paper/2016/hash/9d2682367c3935defcb1f9e247a97c0d-Abstract.html}
  {Equality of opportunity in supervised learning}.
\newblock In \emph{Advances in Neural Information Processing Systems 29: Annual
  Conference on Neural Information Processing Systems 2016, December 5-10,
  2016, Barcelona, Spain}, pages 3315--3323.

\bibitem[{Harris et~al.(2020)Harris, Millman, van~der Walt, Gommers, Virtanen,
  Cournapeau, Wieser, Taylor, Berg, Smith, Kern, Picus, Hoyer, van Kerkwijk,
  Brett, Haldane, del R{\'{i}}o, Wiebe, Peterson, G{\'{e}}rard-Marchant,
  Sheppard, Reddy, Weckesser, Abbasi, Gohlke, and Oliphant}]{harris2020array}
Charles~R. Harris, K.~Jarrod Millman, St{\'{e}}fan~J. van~der Walt, Ralf
  Gommers, Pauli Virtanen, David Cournapeau, Eric Wieser, Julian Taylor,
  Sebastian Berg, Nathaniel~J. Smith, Robert Kern, Matti Picus, Stephan Hoyer,
  Marten~H. van Kerkwijk, Matthew Brett, Allan Haldane, Jaime~Fern{\'{a}}ndez
  del R{\'{i}}o, Mark Wiebe, Pearu Peterson, Pierre G{\'{e}}rard-Marchant,
  Kevin Sheppard, Tyler Reddy, Warren Weckesser, Hameer Abbasi, Christoph
  Gohlke, and Travis~E. Oliphant. 2020.
\newblock \href {https://doi.org/10.1038/s41586-020-2649-2} {Array programming
  with {NumPy}}.
\newblock \emph{Nature}, 585(7825):357--362.

\bibitem[{He et~al.(2016)He, Zhang, Ren, and Sun}]{he2016deep}
Kaiming He, Xiangyu Zhang, Shaoqing Ren, and Jian Sun. 2016.
\newblock Deep residual learning for image recognition.
\newblock In \emph{Proceedings of the IEEE conference on computer vision and
  pattern recognition}, pages 770--778.

\bibitem[{Huang et~al.(2020)Huang, Xing, Dernoncourt, and
  Paul}]{huang-etal-2020-multilingual}
Xiaolei Huang, Linzi Xing, Franck Dernoncourt, and Michael~J. Paul. 2020.
\newblock \href {https://aclanthology.org/2020.lrec-1.180} {Multilingual
  {T}witter corpus and baselines for evaluating demographic bias in hate speech
  recognition}.
\newblock In \emph{Proceedings of the Twelfth Language Resources and Evaluation
  Conference}, pages 1440--1448, Marseille, France. European Language Resources
  Association.

\bibitem[{Iosifidis et~al.(2019)Iosifidis, Fetahu, and
  Ntoutsi}]{iosifidis2019fae}
Vasileios Iosifidis, Besnik Fetahu, and Eirini Ntoutsi. 2019.
\newblock Fae: A fairness-aware ensemble framework.
\newblock In \emph{2019 IEEE International Conference on Big Data (Big Data)},
  pages 1375--1380. IEEE.

\bibitem[{Kamiran and Calders(2009)}]{kamiran2009classifying}
Faisal Kamiran and Toon Calders. 2009.
\newblock Classifying without discriminating.
\newblock In \emph{2009 2nd international conference on computer, control and
  communication}, pages 1--6. IEEE.

\bibitem[{Kamiran and Calders(2012)}]{kamiran2012data}
Faisal Kamiran and Toon Calders. 2012.
\newblock Data preprocessing techniques for classification without
  discrimination.
\newblock \emph{Knowledge and Information Systems}, 33(1):1--33.

\bibitem[{Kang et~al.(2022)Kang, Xie, Wu, Maciejewski, and
  Tong}]{DBLP:conf/bigdataconf/KangXWMT22}
Jian Kang, Tiankai Xie, Xintao Wu, Ross Maciejewski, and Hanghang Tong. 2022.
\newblock \href {https://doi.org/10.1109/BIGDATA55660.2022.10020588} {Infofair:
  Information-theoretic intersectional fairness}.
\newblock In \emph{{IEEE} International Conference on Big Data, Big Data 2022,
  Osaka, Japan, December 17-20, 2022}, pages 1455--1464. {IEEE}.

\bibitem[{Kearns et~al.(2018)Kearns, Neel, Roth, and
  Wu}]{DBLP:conf/icml/KearnsNRW18}
Michael~J. Kearns, Seth Neel, Aaron Roth, and Zhiwei~Steven Wu. 2018.
\newblock \href {http://proceedings.mlr.press/v80/kearns18a.html} {Preventing
  fairness gerrymandering: Auditing and learning for subgroup fairness}.
\newblock In \emph{Proceedings of the 35th International Conference on Machine
  Learning, {ICML} 2018, Stockholmsm{\"{a}}ssan, Stockholm, Sweden, July 10-15,
  2018}, volume~80 of \emph{Proceedings of Machine Learning Research}, pages
  2569--2577. {PMLR}.

\bibitem[{Kingma and Ba(2015)}]{DBLP:journals/corr/KingmaB14}
Diederik~P. Kingma and Jimmy Ba. 2015.
\newblock \href {http://arxiv.org/abs/1412.6980} {Adam: {A} method for
  stochastic optimization}.
\newblock In \emph{3rd International Conference on Learning Representations,
  {ICLR} 2015, San Diego, CA, USA, May 7-9, 2015, Conference Track
  Proceedings}.

\bibitem[{Kirk et~al.(2021)Kirk, Jun, Volpin, Iqbal, Benussi, Dreyer,
  Shtedritski, and Asano}]{DBLP:conf/nips/KirkJVIBDSA21}
Hannah~Rose Kirk, Yennie Jun, Filippo Volpin, Haider Iqbal, Elias Benussi,
  Fr{\'{e}}d{\'{e}}ric~A. Dreyer, Aleksandar Shtedritski, and Yuki~M. Asano.
  2021.
\newblock \href
  {https://proceedings.neurips.cc/paper/2021/hash/1531beb762df4029513ebf9295e0d34f-Abstract.html}
  {Bias out-of-the-box: An empirical analysis of intersectional occupational
  biases in popular generative language models}.
\newblock In \emph{Advances in Neural Information Processing Systems 34: Annual
  Conference on Neural Information Processing Systems 2021, NeurIPS 2021,
  December 6-14, 2021, virtual}, pages 2611--2624.

\bibitem[{Lalor et~al.(2022)Lalor, Yang, Smith, Forsgren, and
  Abbasi}]{DBLP:conf/naacl/LalorYSFA22}
John Lalor, Yi~Yang, Kendall Smith, Nicole Forsgren, and Ahmed Abbasi. 2022.
\newblock \href {https://doi.org/10.18653/v1/2022.naacl-main.263} {Benchmarking
  intersectional biases in {NLP}}.
\newblock In \emph{Proceedings of the 2022 Conference of the North American
  Chapter of the Association for Computational Linguistics: Human Language
  Technologies, {NAACL} 2022, Seattle, WA, United States, July 10-15, 2022},
  pages 3598--3609. Association for Computational Linguistics.

\bibitem[{Li et~al.(2018)Li, Baldwin, and Cohn}]{li-etal-2018-towards}
Yitong Li, Timothy Baldwin, and Trevor Cohn. 2018.
\newblock \href {https://doi.org/10.18653/v1/P18-2005} {Towards robust and
  privacy-preserving text representations}.
\newblock In \emph{Proceedings of the 56th Annual Meeting of the Association
  for Computational Linguistics (Volume 2: Short Papers)}, pages 25--30,
  Melbourne, Australia. Association for Computational Linguistics.

\bibitem[{Liu et~al.(2015)Liu, Luo, Wang, and Tang}]{DBLP:conf/iccv/LiuLWT15}
Ziwei Liu, Ping Luo, Xiaogang Wang, and Xiaoou Tang. 2015.
\newblock \href {https://doi.org/10.1109/ICCV.2015.425} {Deep learning face
  attributes in the wild}.
\newblock In \emph{2015 {IEEE} International Conference on Computer Vision,
  {ICCV} 2015, Santiago, Chile, December 7-13, 2015}, pages 3730--3738. {IEEE}
  Computer Society.

\bibitem[{Lohaus et~al.(2020)Lohaus, Perrot, and Von~Luxburg}]{lohaus2020too}
Michael Lohaus, Micha{\"e}l Perrot, and Ulrike Von~Luxburg. 2020.
\newblock Too relaxed to be fair.
\newblock In \emph{International Conference on Machine Learning}, pages
  6360--6369. PMLR.

\bibitem[{Maheshwari et~al.(2023)Maheshwari, Bellet, Denis, and
  Keller}]{maheshwari-etal-2023-fair}
Gaurav Maheshwari, Aur{\'e}lien Bellet, Pascal Denis, and Mikaela Keller. 2023.
\newblock \href {https://doi.org/10.18653/v1/2023.emnlp-main.558} {Fair without
  leveling down: A new intersectional fairness definition}.
\newblock In \emph{Proceedings of the 2023 Conference on Empirical Methods in
  Natural Language Processing}, pages 9018--9032, Singapore. Association for
  Computational Linguistics.

\bibitem[{Maheshwari and Perrot(2022)}]{DBLP:journals/corr/abs-2206-10923}
Gaurav Maheshwari and Micha{\"{e}}l Perrot. 2022.
\newblock \href {https://doi.org/10.48550/arXiv.2206.10923} {Fairgrad: Fairness
  aware gradient descent}.
\newblock \emph{CoRR}, abs/2206.10923.

\bibitem[{Metcalf and Crawford(2016)}]{metcalf2016human}
Jacob Metcalf and Kate Crawford. 2016.
\newblock Where are human subjects in big data research? the emerging ethics
  divide.
\newblock \emph{Big Data \& Society}, 3(1):2053951716650211.

\bibitem[{Mittelstadt et~al.(2023)Mittelstadt, Wachter, and
  Russell}]{DBLP:journals/corr/abs-2302-02404}
Brent~D. Mittelstadt, Sandra Wachter, and Chris Russell. 2023.
\newblock \href {https://doi.org/10.48550/arXiv.2302.02404} {The unfairness of
  fair machine learning: Levelling down and strict egalitarianism by default}.
\newblock \emph{CoRR}, abs/2302.02404.

\bibitem[{Morina et~al.(2019)Morina, Oliinyk, Waton, Marusic, and
  Georgatzis}]{DBLP:journals/corr/abs-1911-01468}
Giulio Morina, Viktoriia Oliinyk, Julian Waton, Ines Marusic, and Konstantinos
  Georgatzis. 2019.
\newblock \href {http://arxiv.org/abs/1911.01468} {Auditing and achieving
  intersectional fairness in classification problems}.
\newblock \emph{CoRR}, abs/1911.01468.

\bibitem[{Paszke et~al.(2019)Paszke, Gross, Massa, Lerer, Bradbury, Chanan,
  Killeen, Lin, Gimelshein, Antiga, Desmaison, K{\"{o}}pf, Yang, DeVito,
  Raison, Tejani, Chilamkurthy, Steiner, Fang, Bai, and
  Chintala}]{DBLP:conf/nips/PaszkeGMLBCKLGA19}
Adam Paszke, Sam Gross, Francisco Massa, Adam Lerer, James Bradbury, Gregory
  Chanan, Trevor Killeen, Zeming Lin, Natalia Gimelshein, Luca Antiga, Alban
  Desmaison, Andreas K{\"{o}}pf, Edward~Z. Yang, Zachary DeVito, Martin Raison,
  Alykhan Tejani, Sasank Chilamkurthy, Benoit Steiner, Lu~Fang, Junjie Bai, and
  Soumith Chintala. 2019.
\newblock \href
  {https://proceedings.neurips.cc/paper/2019/hash/bdbca288fee7f92f2bfa9f7012727740-Abstract.html}
  {Pytorch: An imperative style, high-performance deep learning library}.
\newblock In \emph{Advances in Neural Information Processing Systems 32: Annual
  Conference on Neural Information Processing Systems 2019, NeurIPS 2019,
  December 8-14, 2019, Vancouver, BC, Canada}, pages 8024--8035.

\bibitem[{Pedregosa et~al.(2011)Pedregosa, Varoquaux, Gramfort, Michel,
  Thirion, Grisel, Blondel, Prettenhofer, Weiss, Dubourg, Vanderplas, Passos,
  Cournapeau, Brucher, Perrot, and Duchesnay}]{scikit-learn}
F.~Pedregosa, G.~Varoquaux, A.~Gramfort, V.~Michel, B.~Thirion, O.~Grisel,
  M.~Blondel, P.~Prettenhofer, R.~Weiss, V.~Dubourg, J.~Vanderplas, A.~Passos,
  D.~Cournapeau, M.~Brucher, M.~Perrot, and E.~Duchesnay. 2011.
\newblock Scikit-learn: Machine learning in {P}ython.
\newblock \emph{Journal of Machine Learning Research}, 12:2825--2830.

\bibitem[{Ravfogel et~al.(2020)Ravfogel, Elazar, Gonen, Twiton, and
  Goldberg}]{DBLP:conf/acl/RavfogelEGTG20}
Shauli Ravfogel, Yanai Elazar, Hila Gonen, Michael Twiton, and Yoav Goldberg.
  2020.
\newblock \href {https://doi.org/10.18653/v1/2020.acl-main.647} {Null it out:
  Guarding protected attributes by iterative nullspace projection}.
\newblock In \emph{Proceedings of the 58th Annual Meeting of the Association
  for Computational Linguistics, {ACL} 2020, Online, July 5-10, 2020}, pages
  7237--7256. Association for Computational Linguistics.

\bibitem[{Thanh-Tung and Tran(2020)}]{thanh2020catastrophic}
Hoang Thanh-Tung and Truyen Tran. 2020.
\newblock Catastrophic forgetting and mode collapse in gans.
\newblock In \emph{2020 international joint conference on neural networks
  (ijcnn)}, pages 1--10. IEEE.

\bibitem[{Weidinger et~al.(2021)Weidinger, Mellor, Rauh, Griffin, Uesato,
  Huang, Cheng, Glaese, Balle, Kasirzadeh et~al.}]{weidinger2021ethical}
Laura Weidinger, John Mellor, Maribeth Rauh, Conor Griffin, Jonathan Uesato,
  Po-Sen Huang, Myra Cheng, Mia Glaese, Borja Balle, Atoosa Kasirzadeh, et~al.
  2021.
\newblock Ethical and social risks of harm from language models.
\newblock \emph{arXiv preprint arXiv:2112.04359}.

\bibitem[{Yang et~al.(2020)Yang, Cisse, and Koyejo}]{DBLP:conf/nips/YangCK20}
Forest Yang, Mouhamadou Cisse, and Oluwasanmi Koyejo. 2020.
\newblock \href
  {https://proceedings.neurips.cc/paper/2020/hash/29c0605a3bab4229e46723f89cf59d83-Abstract.html}
  {Fairness with overlapping groups; a probabilistic perspective}.
\newblock In \emph{Advances in Neural Information Processing Systems 33: Annual
  Conference on Neural Information Processing Systems 2020, NeurIPS 2020,
  December 6-12, 2020, virtual}.

\bibitem[{Zafar et~al.(2017)Zafar, Valera, Rogriguez, and
  Gummadi}]{zafar2017fairnessb}
Muhammad~Bilal Zafar, Isabel Valera, Manuel~Gomez Rogriguez, and Krishna~P
  Gummadi. 2017.
\newblock Fairness constraints: Mechanisms for fair classification.
\newblock In \emph{Artificial Intelligence and Statistics}, pages 962--970.
  PMLR.

\bibitem[{Zietlow et~al.(2022)Zietlow, Lohaus, Balakrishnan, Kleindessner,
  Locatello, Sch{\"{o}}lkopf, and Russell}]{DBLP:conf/cvpr/ZietlowLBKLS022}
Dominik Zietlow, Michael Lohaus, Guha Balakrishnan, Matth{\"{a}}us
  Kleindessner, Francesco Locatello, Bernhard Sch{\"{o}}lkopf, and Chris
  Russell. 2022.
\newblock \href {https://doi.org/10.1109/CVPR52688.2022.01016} {Leveling down
  in computer vision: Pareto inefficiencies in fair deep classifiers}.
\newblock In \emph{{IEEE/CVF} Conference on Computer Vision and Pattern
  Recognition, {CVPR} 2022, New Orleans, LA, USA, June 18-24, 2022}, pages
  10400--10411. {IEEE}.

\end{thebibliography}
\bibliographystyle{acl_natbib}

\clearpage
\appendix

\section{Additional Related Work}
\label{app:related-work}
With ML rapidly automating several key aspects of decsion making the potential for harm has sparked calls for greater accountability and transparency by researchers~\citep{weidinger2021ethical, burrell2016machine, metcalf2016human}, government agencies~\citep{european2018communication,barocas2017big} and NGOs~\citep{buchanan2012ethical}. This has peaked interest in fairness, with researchers responding in two primary ways: (i) Capturing and defining unfairness by proposing new metrics and evaluation suites, and (ii) developing mechanism to mitigate the unfairness.

The most prevalent approach to assessing intersectional unfairness involves comparing subgroup performances either with the overall population, as in subgroup fairness~\citep{DBLP:conf/icml/KearnsNRW18}, or with the best and worst performing subgroups, as in Differential Fairness~\citep{DBLP:conf/icde/FouldsIKP20}. Recent observations by~\citep{maheshwari-etal-2023-fair, DBLP:journals/corr/abs-2302-02404} suggest that solely focusing on relative performance among subgroups, without considering absolute performance, can lead to a phenomenon known as "leveling down." To address this, they recommend a hybrid metric, $\IFa{\alpha}$, which combines relative and absolute performance measures. Further details about these metrics are provided in the Appendix~\ref{app:fairness-measure}.

Mitigation techniques can be typically categorized into three groups: (i) pre-processing, involving modifications at the dataset level~\citep{kamiran2012data, feldman2015certifying, calmon2017optimized}; (ii) post-processing, which adjusts the outputs of pre-trained models that may exhibit biases~\citep{iosifidis2019fae, chzhen2019leveraging}; and (iii) in-processing, entailing alterations to the training process and the model itself to enhance fairness~\citep{cotter2019two, lohaus2020too, calders2010three}.

% In terms of approaches w

In terms of approaches that specifically optimize intersectional fairness,~\citet{DBLP:conf/icde/FouldsIKP20} introduced an in-processing technique that incorporates a fairness regularizer into the loss function, balancing fairness and accuracy. Conversely,~\citet{DBLP:journals/corr/abs-1911-01468} suggests a post-processing mechanism that adjusts the threshold of the classifier and randomizes predictions for each subgroup independently. InfoFair~\citep{DBLP:conf/bigdataconf/KangXWMT22} adopts a distinct approach by minimizing mutual information between predictions and sensitive attributes. Recently, research has begun to explore the phenomenon of "leveling down" in fairness.~\citet{maheshwari-etal-2023-fair, DBLP:journals/corr/abs-2302-02404} argue that the strictly egalitarian perspective of current fairness measures contributes to this phenomenon. Meanwhile,~\citet{DBLP:conf/cvpr/ZietlowLBKLS022} demonstrates leveling down in computer vision contexts and introduces an adaptive augmented sampling strategy using generative adversarial networks~\citep{DBLP:journals/corr/GoodfellowPMXWOCB14} and SMOTE~\citep{DBLP:journals/jair/ChawlaBHK02}. Our work aligns with these developments; however, we propose a modality-independent technique that effectively leverages the intrinsic structure of intersectionality.

\section{Background on Maximum Mean Discrepancy}
\label{app:mmd}

Maximum Mean Discrepancy is an non-parametric kernel-based divergence used to assess the similarity between distributions. In a nutshell, it involves identifying an embedding function that, given two distributions $\mathcal{P}$ and $\mathcal{Q}$, yields larger values for samples drawn from $\mathcal{P}$ and smaller values for those from $\mathcal{Q}$. The difference in the mean value of this function for samples drawn from these two distributions provides an estimate of their similarity.

In this work, following the footsteps of~\citet{DBLP:journals/jmlr/GrettonBRSS12}, we use unit balls in characteristic reproducing kernel Hilbert spaces as the function class. Intuitively, the idea is to use the kernel trick to compute the differences in all moments of two distributions and then average the result. Formally, the MMD between two distributions $\mathcal{P}$ and $\mathcal{Q}$ is:
\begin{align*}
    &MMD^2(\mathcal{P}, \mathcal{Q}) \\
    = &\quad \textstyle{\sup_{\left \| \Psi \right \|_H \leq 1 }}|E_{Z \sim \mathcal{P}}[\Psi(Z)]  - E_{Z' \sim \mathcal{Q}}[\Psi(Z')]| \\
    = &\quad E_{Z \sim \mathcal{P}} [k(Z,Z)]
    - 2E_{Z \sim \mathcal{P}, Z' \sim \mathcal{Q}} [k(Z,Z')] \\
    &+ \quad E_{Z' \sim \mathcal{Q}} [k(Z',Z')]\\
\end{align*}

Here, $k$ is the kernel derived from  $\left \| \cdot \right \|_{H}$, the norm associated with corresponding Reproducing Kernel Hilbert Space $H$. In practice, we generally do not have access to true distributions but only samples, and thus the above equation is approximated as:

\begin{align*}
    &MMD^2(S_{z}, S_{z'}) = \textstyle\frac{1}{m(m-1)}\big[ \sum_{i} \sum_{j \neq i} k(z_i, z_j)\big. & \\
    & \big. +\textstyle\sum_{i} \sum_{j \neq i} k(z'_i, z'_j)\big] + \frac{1}{m^2}\sum_{i} \sum_{j} k(z_i, z'_j) &
\end{align*}
where $S_{z}$ (resp. $S_{z'}$) is a set of m samples drawn from $\mathcal{P}$ (resp. $\mathcal{Q}$). In this work, we use the radial basis function kernel $k: (z,z') \mapsto \exp(\left \| z-z' \right \|^2 / 2\sigma^2)$ where $\sigma$ is the free parameter. In summary, MMD provides a simple way to compute the similarity between two distributions by using samples drawn from those distributions.

\section{Algorithm}
\label{app:algorithm}

The procedure to train our generative models is summarized in Algorithm~\ref{alg:generator}.

\begin{algorithm}[t]
% \algsetup{linenosize=\tiny}
% \scriptsize
\caption{Training the Generative Models}
\label{alg:datagenerations}
\textbf{Input}: Groups $\spaceG$, Dataset $\setT$, % initial generative model $gen_{\theta}$,
batch size $b$, number of iterations $l$ and batch size $b$ \\
\textbf{Output}: $K$ trained generative models $\{gen_{\theta,k}\}_{k=1}^K$ capable of generating data for each label $k$ % The following space should be kept as it prevents weird behaviours from the package

\begin{algorithmic}[1] %[1] enables line numbers
\FOR{\_ in $l$}
% \FOR{Group $\mathbf{g}$  in $\spaceG$}
\STATE Randomly sample a group $\mathbf{g}$ from $\spaceG$
\FOR{ $k$ in $K$}
\STATE $S_{\mathbf{g},k}$ $\leftarrow$  Sample $b$ examples from $\mathcal{T}_{\mathbf{g}|Y=k}$
\STATE $S_{\mathbf{g}^{\backslash i},k}$ $\leftarrow$ Sample $b$ examples from $\mathcal{T}_{\mathbf{g}^{\backslash i}|Y=k} \: \forall i\in\{1,\dots, p\}$
% \STATE Embed each set of example using a pre-trained model
\STATE $S_{gen}$ $\leftarrow$ Sample $b$ examples  from $gen_{\theta,k}(\setT, \mathbf{g})$ % such as the one in Equation~\ref{data-gen:simple-gen-fn}
\STATE Compute the MMD loss using these examples as stated in Equation~\ref{data-gen:loss-mdd-sample}
\STATE Backpropagate this loss to update the parameters of the model $gen_{\theta,k}$
% (\setT, \mathbf{g})$
\ENDFOR
% \ENDFOR
\ENDFOR
  % \IF{exit condition is true}
  %   \STATE return all trained models of the form  $Gen_{\theta}{}_{m}$
  % \ELSE
  %   \STATE Repeat Step 1
  % \ENDIF

\end{algorithmic}
\label{alg:generator}
\end{algorithm}

\section{Experiments}

\subsection{Datasets}
\label{app:datasets}

We benchmark our proposed generative approach over four datasets, and employ a similar setup as proposed by~\citep{maheshwari-etal-2023-fair}. Note that all the datasets we experiment with are publicly available and can be used for research purpose. 

\begin{itemize}

\item ~\textit{CelebA}~\citep{DBLP:conf/iccv/LiuLWT15}: It is composed of 202, 599 images of human faces. Additionally, each image is annotated with  40 binary attributes, such as  ‘eye glasses’, ‘bangs’, and ‘mustaches’. In our experiments, we set ‘sex’, ‘Young’, ‘Attractive’, and ‘Pale Skin’ attributes as the sensitive axis for the images and ‘Smiling’ as the class label. We split the dataset into 80\% training of which 20\% is used as validation, and the remaining 20\% test split.

\item \textit{Twitter Hate Speech}~\citep{huang-etal-2020-multilingual}: The dataset consists of tweets annotated with four race, age, gender, and country, We use the same pre-processing steps as employed by~\citep{maheshwari-etal-2023-fair}, including binarizing the sensitive attributes, and focusing on English subset. After pre-processing, our train, validation and test sets consists of $22,818$, $4,512$, and $5,032$ tweets respectively.

\item \textit{Psychometric dataset}~\citep{DBLP:conf/emnlp/AbbasiDLNSY21}: The dataset consists of $8,502$ text responses alongside numerical scores provided by the physicians over several psychometric dimensions. Each response is also associated with four sensitive attributes, namely gender, race, and age. We focus on:
\begin{itemize}
    \item \textit{Numeracy} which reflects the numerical comprehension ability of the patient.
    \item \textit{Anxiety} reflects the level of anxiety as described by the adult.
\end{itemize}

We use same pre-processing as~\cite{DBLP:conf/naacl/LalorYSFA22} including binarizing the score. We use the same splitting procedure as described for CelebA dataset.

\end{itemize}

\subsection{Hyperparameters}
\label{app:hyperparameters}

In all our experiments, we utilized an Intel Xeon CPU. Training a generative model on this platform typically takes about 15 minutes, whereas our fairness-accuracy experiments generally required about 30 minutes. For ease of replication, we will include the PyTorch model description in the README file accompanying the source code. All experiments were conducted using five different seeds: 10, 20, 30, 40, and 50. For the \Adv{} approach, the $\lambda$ parameter, which indicates the weight assigned to the adversarial branch, was set to the following values: 0.25, 0.5, 1.0, 5.0, 10.0, 50.0, 100.0. Similarly, for \FairMixup{}, the mixup regularizer was assigned values of 0.25, 0.5, 1.0, 5.0, 10.0, 50.0, 100.0. For all other approaches, we used the default settings from the respective authors' codebases. The selection of optimal hyperparameters followed the procedure outlined in~\citep{DBLP:journals/corr/abs-2206-10923}. In every experiment, we fixed the value of $k$ at 0.03.

\subsection{Fairness Definitions}
\label{app:fairness-measure}

In this work, we utilize two fairness definitions specifically formulated to assess intersectional fairness. Both definitions depend on group-wise performance measures, denoted as $m$, which can take various forms, including Accuracy, True Positive Rate, and False Positive Rate. We focus on False Positive Rate for which the corresponding $m$ is:

\begin{align*}
    m(h_{\theta}, \setT_{\mathbf{g}}) = 1 - P(h_{\theta}(x)=0|(x,y) \in \setT_{\mathbf{g}}, y=1)
    % \: \forall (x,y)\: \in \setT_{\mathbf{g}}
\end{align*}

\begin{itemize}
    \item Differential Fairness: A model, denoted by $h_{\theta}$, is considered to be  $\epsilon$-differentially fair (\DF) wrt  $m$, if
% with respect to a group-wise performance measure $m$, if
    \begin{align*}
        \DF(h_{\theta}, m) \equiv \underset{\mathbf{g}, \mathbf{g'} \in \spaceG}{\max}\: \: \log \frac{m(\fh[_\theta]{}, \setT_{\mathbf{g}})}{m(\fh[_\theta]{}, \setT_{\mathbf{g'}})} \leq \epsilon.
    \end{align*}

    \item $\alpha$-Intersectional Framework: A model $h_{\theta}$ is said to be $(\alpha, \gamma)$-intersectionally fair ($\IFa{\alpha}$) with respect to  $m$, if $$\IFa{\alpha}(h_{\theta}, m) \equiv  \underset{\mathbf{g}, \mathbf{g}' \in \spaceG}{\max} I_{\alpha}(\mathbf{g}, \mathbf{g}', h_{\theta}, m) \leq \gamma. $$

    where $\mathbf{g}^w = {\argmin_{\mathbf{g} \in \spaceG}}\: m(h_{\theta}, \setT_{\mathbf{g}})$ and $\mathbf{g}^b = {\argmax_{\mathbf{g} \in \spaceG}}\: m(h_{\theta}, \setT_{\mathbf{g}})$.  Here $I_{\alpha}(\mathbf{g}, \mathbf{g}', h_{\theta}, m)$ is defined as:
    \begin{equation}
        I_{\alpha}(\mathbf{g}, \mathbf{g}', h_{\theta}, m) = \alpha\Delta_{abs} + (1-\alpha)\Delta_{rel},
    \end{equation}
     where $\alpha \in [0,1]$ and
    \begin{align*}
        \Delta_{abs} &= \max\left(1-m(\fh[_\theta]{}, \setT_{\mathbf{g}}),1-m(\fh[_\theta]{}, \setT_{\mathbf{g}'})\right), \\
        \Delta_{rel} &= \frac{1 - \max\left(m(\fh[_\theta]{}, \setT_{\mathbf{g}}),m(\fh[_\theta]{}, \setT_{\mathbf{g}'})\right)}{1 - \min\left(m(\fh[_\theta]{}, \setT_{\mathbf{g}}),m(\fh[_\theta]{}, \setT_{\mathbf{g}'})\right)}.
    \end{align*}

\end{itemize}

\subsection{ Results}
\label{app:extended-results}

\begin{figure*}% <---
\centering

   \begin{subfigure}{0.45\textwidth}
       \includegraphics[width=\linewidth]{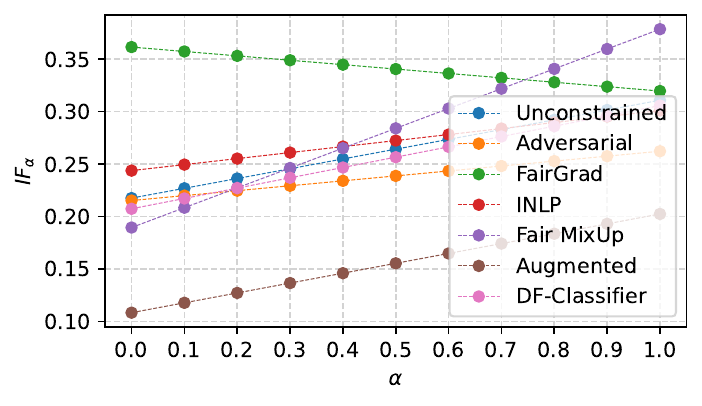}
       \caption{CelebA FPR}
   \end{subfigure}
\hfill % <--- 
   \begin{subfigure}{0.45\textwidth}
       \includegraphics[width=\linewidth]{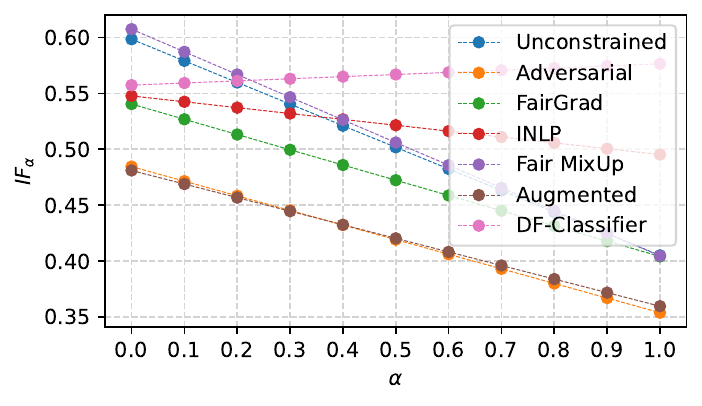}
       \caption{Numeracy FPR}
   \end{subfigure}
\hfill % <--- 
   \begin{subfigure}{0.45\textwidth}
      \includegraphics[width=\linewidth]{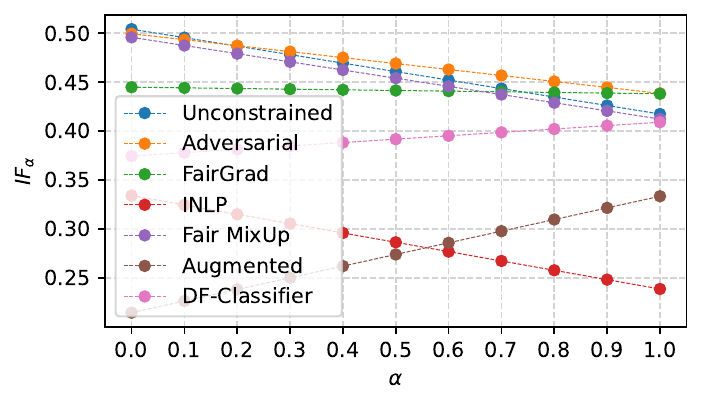}
       \caption{Twitter Hate Speech FPR}
   \end{subfigure}
\hfill % <--- 
   \begin{subfigure}{0.45\textwidth}
       \includegraphics[width=\linewidth]{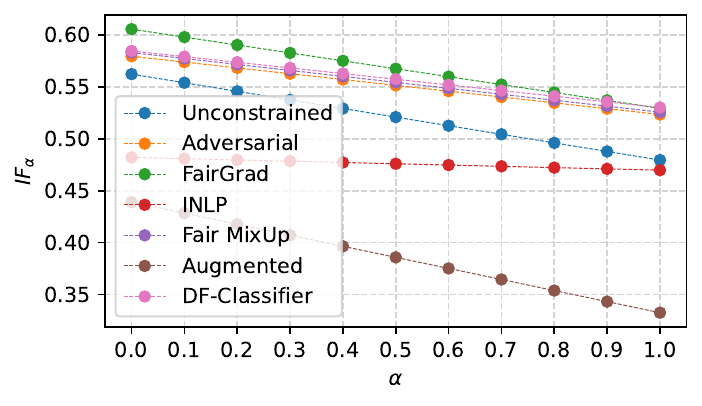}
       \caption{Anxiety FPR}
   \end{subfigure}

   \caption{Value of $\IFa{\alpha}$ on the test set of various datasets by varying $\alpha \in [0,1]$.}
   \label{app:fig:overall-tradeoff}
\end{figure*}

We detail the additional experiments over the \textit{CelebA} and \textit{Numeracy} datasets. Table~\ref{tab:fairness-accuracy-app} shows results for fixed value of $\alpha$. While Figure~\ref{app:fig:overall-tradeoff} plot the trade-off between relative and absolute performance over groups by varying $\alpha$ for all the datasets.

\begin{table*}[h!]
    \centering

    \begin{subtable}{\linewidth}
    \vspace*{0.5 cm}
        \centering
        \adjustbox{max width=\linewidth}{
        \begin{tabular}{@{}lccccc@{}}
            \toprule
                Method          & BA           & Best Off    & Worst Off   & \DF{}       & $\IFa{0.5}$ \\ \midrule
                \Uncon          & 0.81 + 0.0  & 0.06 + 0.02 & 0.34 + 0.01 & 0.35 +/- 0.38 & 0.26 +/- 0.04\\
                \Adv            & 0.81 + 0.01  & 0.05 + 0.01 & 0.3 + 0.03 & 0.31 +/- 0.19 & 0.24 +/- 0.03\\
                \FairGrad       & 0.76 + 0.0  & 0.1 + 0.01 & 0.35 + 0.04 & 0.33 +/- 0.12 & 0.34 +/- 0.02 \\
                \INLP           & 0.81 + 0.01  & 0.07 + 0.01 & 0.35 + 0.03 & 0.36 +/- 0.16 & 0.27 +/- 0.01\\
                \FairMixup      & 0.81 + 0.0  & 0.06 + 0.0 & 0.4 + 0.07 & 0.45 +/- 0.19 & 0.28 +/- 0.02 \\
                \CDF            & 0.82 + 0.0  & 0.06 + 0.02 & 0.34 + 0.03 & 0.35 +/- 0.33 & 0.26 +/- 0.05 \\
                \UnconA         & 0.76 + 0.01  & 0.02 + 0.0 & 0.21 + 0.03 & 0.22 +/- 0.21 & 0.16 +/- 0.01 \\\bottomrule

\end{tabular}}

        \caption{\label{app-tab:celeba-main} Results on CelebA}
    \end{subtable}

     \begin{subtable}{\linewidth}
    \vspace*{0.5 cm}
        \centering
        \adjustbox{max width=\linewidth}{
        \begin{tabular}{@{}lccccc@{}}
            \toprule
                Method          & BA           & Best Off    & Worst Off   & \DF{}       & $\IFa{0.5}$ \\ \midrule
                \Uncon          & 0.7 + 0.01  & 0.21 + 0.05 & 0.46 + 0.06 & 0.38 +/- 0.13 & 0.5 +/- 0.06\\
                \Adv            & 0.69 + 0.02  & 0.15 + 0.03 & 0.39 + 0.04 & 0.33 +/- 0.16 & 0.42 +/- 0.05 \\
                \FairGrad       & 0.7 + 0.01  & 0.19 + 0.05 & 0.45 + 0.09 & 0.39 +/- 0.12 & 0.47 +/- 0.06 \\
                \INLP           & 0.69 + 0.0  & 0.23 + 0.02 & 0.52 + 0.02 & 0.47 +/- 0.05 & 0.52 +/- 0.02\\
                \FairMixup      & 0.69 + 0.01  & 0.21 + 0.04 & 0.45 + 0.05 & 0.36 +/- 0.09 & 0.51 +/- 0.04 \\
                \CDF            & 0.68 + 0.01  & 0.29 + 0.06 & 0.61 + 0.11 & 0.6 +/- 0.16 & 0.57 +/- 0.07 \\
                \UnconA         & 0.69 + 0.02  & 0.14 + 0.05 & 0.39 + 0.11 & 0.34 +/- 0.24 & 0.44 +/- 0.07 \\\bottomrule

        \end{tabular}}
        \caption{\label{app-tab:numeracy} Results on Numeracy}
    \end{subtable}

    \caption{Test results on (a) \textit{CelebA}, (b) \textit{Numeracy}. We select hyperparameters based on $\IFa{0.5}$ value. The utility of various approaches is measured by balanced accuracy (BA), whereas fairness is measured by differential fairness ($\DF$) and intersectional fairness ($\IFa{0.5}$) on the False Positive Rate (FPR). For both fairness definitions, lower is better, while for balanced accuracy, higher is better. The Best Off and Worst Off, in both cases lower is better, represents the min FPR and max FPR. Results have been averaged over 5 different runs.}
    \label{tab:fairness-accuracy-app}
    
\end{table*}

\subsection{Alternate Structure}
\label{app:sec:hierarchical_structure}

Recall that in \Anarchist{} approach, our aim is to draw examples from a different parent group set. More specifically, we follow an adversarial approach where we choose parents such that they share no examples with the group.

Formally, for a group $\mathbf{g}$ represented as $(a_1, \dots, a_p)$ , we define adversarial group as $\mathbf{\neg g}$ represented by $(\neg a_1, \dots, \neg a_p)$. Note that, in this experiment we assume $\spaceA_{1}, \dots, \spaceA_{p}$ to be binary discrete-valued. The generative function $gen_{\theta,k}(\setT,\mathbf{g})$, akin to Equation~\ref{data-gen:simple-gen-fn}, is defined as:

\begin{equation}
\label{data-gen:simple-gen-fn-adversarial}
    X_{gen} = \sum^{p}_{i=1}\lambda_i X_{\mathbf{\neg g}^{\backslash i}}
\end{equation}

And the corresponding loss function akin to Equation~\ref{data-gen:loss-mdd-sample} is:

\begin{equation}
\label{data-gen:loss-mdd-sample-adversarial}
\begin{split}
    L_{\mathbf{g},k}(\theta) &= MMD(S_{gen}, S_{\mathbf{g},k}) + \\
      &\:\:\:\:\: \sum^{p}_{i=1}MMD(S_{gen}, S_{\mathbf{\neg g}^{\backslash i},k}),
\end{split}
\end{equation}

\subsection{Tools}

In all our experiments, we utilized Python and its associated machine learning libraries, including Numpy~\citep{harris2020array}, PyTorch~\citep{DBLP:conf/nips/PaszkeGMLBCKLGA19}, and scikit-learn~\citep{scikit-learn}. Additionally, we employed ChatGPT for grammar correction.

\end{document}